\documentclass{article}

    \usepackage[nonatbib,final]{neurips_2022}

\usepackage[utf8]{inputenc} 
\usepackage[T1]{fontenc}    
\usepackage{hyperref}       
\usepackage{url}            
\usepackage{booktabs}       
\usepackage{amsfonts}       
\usepackage{nicefrac}       
\usepackage{microtype}      
\usepackage{xcolor}         
\usepackage{graphicx}
\usepackage{bm}
\usepackage{subcaption}
\usepackage{enumitem}
\usepackage{tikz}
\usepackage{pgfplots}
\usepackage{mathtools}
\usepackage[capitalise]{cleveref}

\definecolor{darkred}{rgb}{0.6, 0.1, 0.05}
\definecolor{blueish}{rgb}{0.0, 0.3, .6}
\newcommand{\ds}[1]{{\color{blueish}#1}} 

\newcommand{\helvetica}{\fontfamily{phv}\selectfont\small\textbf}
\newcommand{\s}{\small}

\everypar{\looseness=-1}
\title{\ds{Spatial Trajectories are First-Class Entities}}

\title{Learning Distributions over Spatial Trajectories}

\title{Representing Spatial Trajectories as Distributions}

\author{%
  D\'idac Sur\'is\\
  Columbia University\\
  \texttt{didac.suris@columbia.edu}\\
   \And
   Carl Vondrick\\
   Columbia University\\
   \texttt{vondrick@cs.columbia.edu}\\
}

\begin{document}

\maketitle

\vspace{-0.2cm}
\begin{abstract}
  We introduce a representation learning framework for spatial trajectories. We represent partial observations of trajectories as probability distributions in a learned latent space, which characterize the uncertainty about unobserved parts of the trajectory.
  Our framework allows us to obtain samples from a trajectory for any continuous point in time---both interpolating and extrapolating. Our flexible approach supports directly modifying specific attributes of a trajectory, such as its pace, as well as combining different partial observations into single representations. Experiments show our method's advantage over baselines in prediction tasks. See \href{http://trajectories.cs.columbia.edu}{\textbf{\textcolor{purple}{\texttt{trajectories.cs.columbia.edu}}}} for video results and code.

\end{abstract}

\vspace{-0.2cm}

\section{Introduction}
\label{sec:introduction}
\vspace{-0.1cm}

The visual world is full of objects moving around in predictable ways. Examples of these \emph{spatial trajectories} include human motion, such as people dancing or exercising; objects moving, such as a ball rolling; trajectories of cars and bicycles; or animal migration patterns. Evidence suggests that the human perceptual system encodes motion into high-level neural codes that represent the motion holistically, going beyond the specific input observations \cite{johansson1973visual}. Humans use this abstract representation for downstream tasks like inferring intention \cite{blakemore2001perception}. Computer vision systems likely need similar mechanisms to encode trajectories and motions into global representations.

Representation learning has been transformative in other domains such as images and text for its ability to obtain high-level representations that reorganize the information in the input, and are better at downstream tasks than the original signals. 
A global representation of trajectories would allow us to evaluate a trajectory at any point in time, even ones not yet observed.
However, modeling trajectories presents a series of challenges for representation learning. First, in real-time scenarios, the future of the trajectory is never observed. Second, temporal and spatial occlusions may impede observing part of a trajectory. Third, trajectories are by nature continuous in time.  
And finally, a trajectory-level metric is usually not well defined and application-dependent. 

We propose a representation learning framework for trajectories that deals with all these challenges in a unified way. Our key contribution is the representation of a partial observation of a trajectory as a probability distribution in a learned latent space, that represents all the possible trajectories the observation could have been sampled from. Our framework's simplicity and generality allows it to be flexible: it does not constrain the input-space metric, accepts observations of different lengths and at any (irregularly sampled) point in time, can be implemented using different families of latent space distributions, and is capable of performing inference-time tasks for which it has not been explicitly trained.
Our experiments on human movement datasets show that our method can accurately predict the past and future of a trajectory segment, as well as the interpolation between two different segments, 
outperforming autoregressive baselines. Additionally, it can do so for any continuous point in time. We also show how we can modify given trajectories by manipulating their representations.

\begin{figure}
  \centering
  \includegraphics[width=\columnwidth]{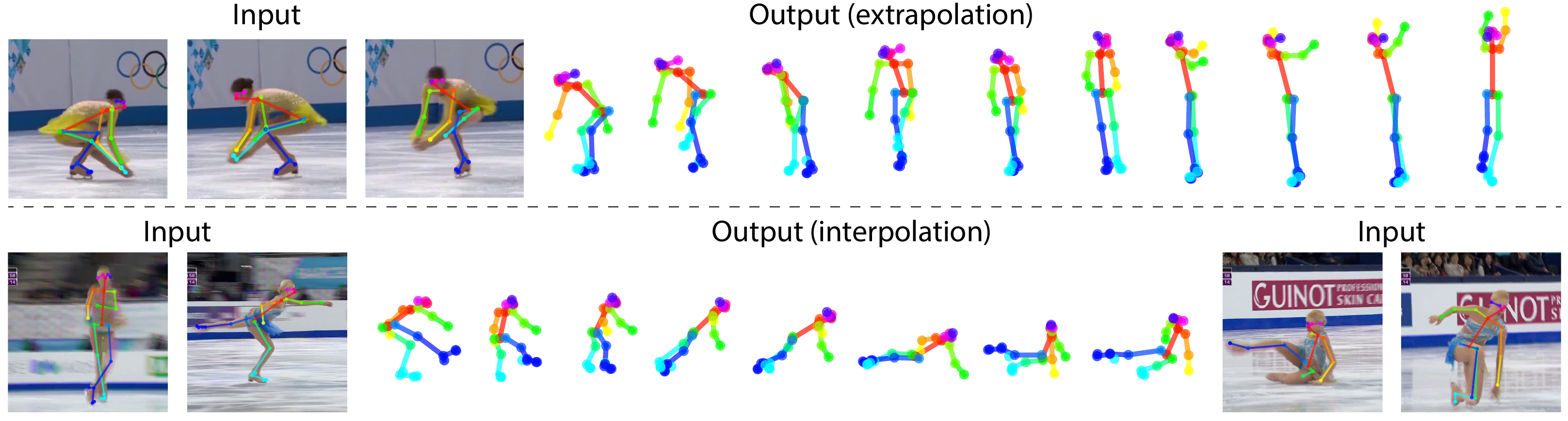}
  \caption{\textbf{Predictions on figure skating data (FisV)}. Our model is capable of predicting the future (top row), past, and interpolation (bottom row) of a trajectory given partial observations, at any continuous time. The inputs to the model are the keypoints in the images. See more examples in \cref{fig:predictions}.}
  \label{fig:teaser}
    \vspace{-0.2cm}
\end{figure}

\section{Method}
\label{sec:method}
\vspace{-0.15cm}

\subsection{Framework, Definitions and Notation}
\vspace{-0.15cm}

The input to our framework is a sequence of samples obtained from a (continuous in time and infinite) spatial trajectory $u$, which we define as the continuous temporal evolution of a set of spatial coordinates. We call each sample a \emph{point} $x$, which lives in the \emph{input space} $\mathbb{R}^K$. We call the sequence of points, together with the times $t$ at which they were sampled, a \emph{segment} $s$, which can be understood as a partial observation of $u$. We define a distance metric $\delta$ between points $x$ in the input space. 

Our goal is to transform these measurements of motion $s$ into a representation $z$ that will be useful for downstream tasks. We define a \emph{latent space} $\mathbb{R}^N$ of \emph{trajectories} $z$. Each $z$ in this space represents the full extent of a trajectory, both in time and in space. 
We use $Q$ to represent probability distributions over trajectories $z$, in the latent space. We define a distance function $D$ between distributions of trajectories, which assumes an underlying distance function $d$ between trajectories $z$. 

We use an encoder $\Theta$ to encode every segment $s$ to a probability distribution $Q(\cdot;s)=\Theta(s)$ over trajectories, where $Q(z;s)$ represents the probability that $s$ was sampled from the trajectory represented by $z$. Additionally, we can decode a trajectory $z$ at a specific time $t$ by using a decoder $\Phi$, obtaining a point $x=\Phi(z,t)$. $\Phi$ takes any continuous $t$ as input. See Fig.~\ref{fig:schematic} for a schematic. 

\subsection{Representation Learning} 
\label{sec:training}
\vspace{-0.15cm}

When observing a segment $s$, one may have some uncertainty about the specific trajectory it was sampled from. For instance, a segment showing a person jumping may correspond to a trajectory that continues with the person falling, or to a trajectory that proceeds with them doing a backflip and landing on their feet, but it will not belong to a trajectory of a person swimming. Therefore, we represent the segment as a distribution over trajectories, where $Q(z;s)$ represents the likelihood of a trajectory given the segment. During training, the goal is to learn this mapping from the input space (segments of trajectories) to the latent space (distributions over trajectories).  

\looseness=-1
Concretely, given two segments $s^a, s^b$ that have been obtained from the same underlying trajectory, we want some $z$ to exist such that its likelihood under the distributions $Q^a$ and $Q^b$ representing each of the segments is high. To encourage this, we train the model to maximize the overlap between the distributions $Q^a$ and $Q^b$. Similarly, we minimize the overlap between (the distribution representations of) segments sampled from different trajectories, under the assumption that no trajectory $z$ exists that contains both segments. Specifically, we minimize a self-supervised triplet loss:

\begin{figure}
  \centering
  \vspace{-0.1cm}
  \includegraphics[width=\columnwidth]{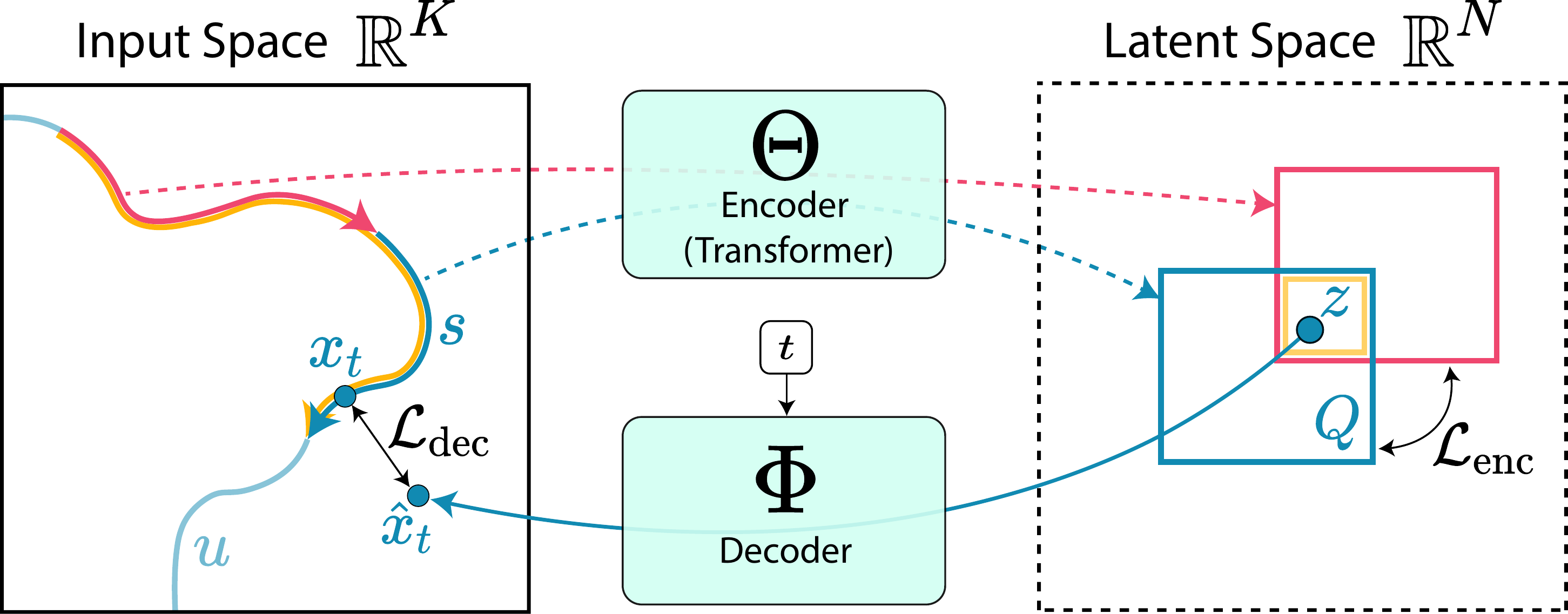}
  \caption{\textbf{Schematic of our framework}. We show the input space $\mathbb{R}^K$, the latent space $\mathbb{R}^N$, and the mappings between the two (encoder $\Theta$ and decoder $\Phi$). A segment $s$ belonging to a trajectory $u$ is encoded into a distribution $Q$, from which a trajectory $z$ is sampled and decoded at a time $t$, to get $\hat{x}_t$.}
  \label{fig:schematic}
   \vspace{-0.2cm}

\end{figure}

\begin{equation}\label{eq:traj}
    \mathcal{L}_{\mathrm{enc}} =\sum_{(i,k^+,k^-)\in\mathcal{T}}\max \left[D\left(Q^i, Q^{k^+}\right)-D\left(Q^i, Q^{k^-}\right)+\alpha, 0\right],
\end{equation}

where $\alpha$ is a margin hyperparameter, and $\mathcal{T}$ is a set of triplets: for every segment $i$ in the dataset, we define several triplets by sampling pairs consisting of a positive segment $k^+$ (such that $(i,k^+)$ is a positive pair) and a negative segment $k^-$ (such that $(i,k^-)$ is a negative pair). 

In addition to learning representations of trajectories, we also wish to be able decode them back to input-space points. To achieve this, we train a decoder $\Phi$ that allows us to obtain the specific value of any trajectory at any continuous time $t$. In order to train the decoder $\Phi$, we sample trajectories $z\sim Q(\cdot;s)$ from each segment representation, and decode them at specific time-steps $t$ which were contained in $s$, obtaining a prediction $\hat{x}_t=\Phi(z,t)$ for which we have ground truth $x_t$. There is no uncertainty in this prediction, as $x_t$ was part of the segment $s$ in the first place; the decoder is only explicitly trained for reconstruction, not extrapolation. We train the decoder via regression, using the \emph{point-wise} distance $\delta$. Note that we never explicitly define a trajectory-level distance in the input space; it is implicitly learned by the model. The reconstruction loss is mathematically 
defined as:
\begin{equation}\label{eq:rec}
    \mathcal{L}_{\mathrm{dec}} = \frac{1}{N}\sum_{i=1}^N\mathbb{E}_{z\sim \Theta(s^i)}\sum_t^{T^i}\delta\left(\Phi(z,t), x_t^i\right),
\end{equation}
where $N$ is the number of segments in the dataset, and $T^i$ is the number of points in segment $s^i$. We minimize \cref{eq:traj,eq:rec} jointly and end-to-end. We implement the encoder $\Theta$ using a Transformer Encoder architecture \cite{transformers}, and the decoder $\Phi$ using a ResNet \cite{resnet}. See Appendix~\ref{apx:details} for more details. 

\subsection{Creating Positive and Negative Pairs}
\label{sec:pairs}

\begin{figure}
  \centering
  \includegraphics[width=\columnwidth]{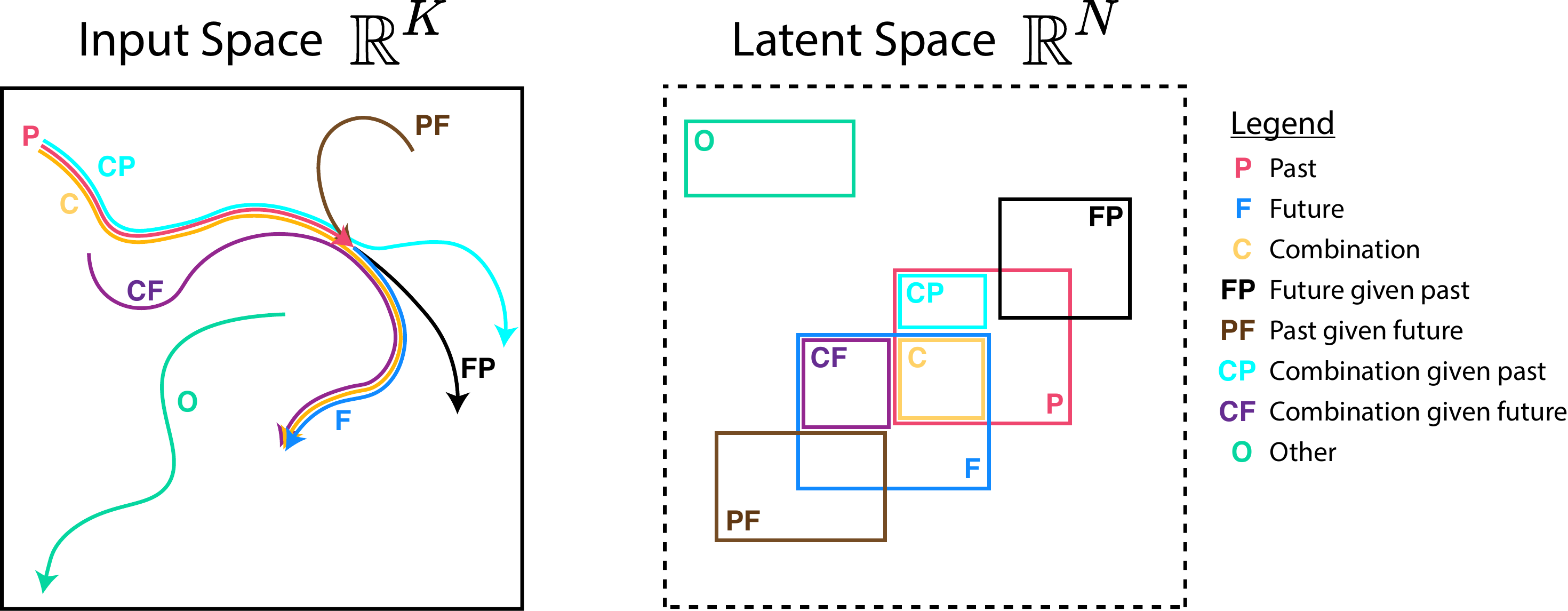}
  \caption{\textbf{Examples of segments}. We illustrate how spatial trajectories (left) are ideally encoded into the latent space (right). The intersection between two segment representations (boxes in the figure) represents the trajectories that contain the two segments.
  ``Future given past'' represents a segment decoded at a future time, from a trajectory sampled from the past representation. It is effectively a sample of a possible future given the past. Other segments are defined similarly. For clarity, we do not show other options like ``past given past'', which would be the same box as past {\color[HTML]{ef476f}{\helvetica P}}. Best viewed in color.}
  \label{fig:latent_space}
\end{figure}

In order to define positives and negative pairs for \cref{eq:traj}, we use the following:

\begin{itemize}[topsep=0pt,itemsep=0ex,partopsep=1ex,parsep=1ex,leftmargin=0.5cm]
\item \textbf{Input-space relationships}. The simplest way is to take segments from the same trajectory as positives and segments from other random trajectories as negatives. The initial segments can have different relationships, such as precedence, containment, or overlap \cite{allen_algebra}. In our experiments, we sample three segments for every trajectory: a \textit{past} segment ({\color[HTML]{ef476f}{\helvetica P}} in Fig.~\ref{fig:latent_space}), a \textit{future} segment ({\color[HTML]{118aff} \helvetica{F}}) whose starting time comes right after the end of the past segment, and a \textit{combination} segment ({\color[HTML]{ffd166} {\helvetica C}}), which contains both the past and the future segments.
\item \textbf{Intersection}. An intersection {\color[HTML]{b36b00} {\helvetica {I}}} of two distributions $Q$ in the latent space will represent all the trajectories that have a high likelihood for both intersected segments. Note that an intersection in the latent space is a union in the input space: the intersection constrains the possible trajectories to those that are consistent simultaneously for the two segments. Similarly, an intersection in the input space (assuming an overlap between segments) is a union in the latent space. In the latent space, the intersection of the past and future segments should be equal to the representation of the combination segment, and therefore the pair ({\color[HTML]{ffd166} {\helvetica C}}, {\color[HTML]{b36b00} {\helvetica {I}}}) is a positive one.
\item \textbf{Re-encoding}. Given a trajectory $z$, we can decode it into any set of times $t$, obtaining a new segment. This segment can be (re-)encoded using $\Theta$, and a representation $Q$ can be obtained for it, resulting in a new positive or negative for other segment representations. For example, when given the past we randomly sample a possible the future, the resulting segment ({\color[HTML]{000000} {\helvetica {FP}}} - \textit{future given past}) will be \textit{different} than the ground truth future, so the pair ({\color[HTML]{118aff} \helvetica{F}}, {\color[HTML]{000000} {\helvetica {FP}}}) will be treated as a negative.
\end{itemize}

We exemplify a combination of these possibilities in Fig.~\ref{fig:latent_space}. In order to determine which pairs of segments are positive, and which are negative, the rule is always the same: if they can belong to the same trajectory they are positives, otherwise they are negatives. For example, looking at Fig.~\ref{fig:latent_space} it is clear that, as discussed above, there is no trajectory that can contain both {\color[HTML]{118aff} \helvetica{F}} and {\color[HTML]{000000} {\helvetica {FP}}}. We list all negative and positive pairs in Appendix~\ref{sec:all_pairs}.

\subsection{Comparing Distributions}
\label{sec:approaches}

\cref{eq:traj} uses the distance function $D$ to compare distributions of trajectories. In this section, we introduce two different ways of designing $D$, resulting in different intuitions about the latent space.

\paragraph{Symmetric Distance} If two segments can belong to the same trajectory, the distributions $Q$ representing each segment should be similar and close to each other (positives), and the (symmetric) distance $D$ between them should be small. For example, in Fig.~\ref{fig:latent_space}, the representations of the past {\color[HTML]{ef476f}{\helvetica P}} and future {\color[HTML]{118aff} \helvetica{F}} segments belonging to the same trajectory are treated as positives.

\paragraph{Conditional} Instead of computing a distance or a similarity, we compute the probability that a segment $s^a$ belongs to the same trajectory as another segment $s^b$. We model this as a conditional probability $P(Q^a|Q^b)$. There are four possibilities: 
\begin{enumerate}[topsep=0pt,itemsep=-1ex,partopsep=1ex,parsep=1ex,leftmargin=0.5cm]
\item $P(Q^a|Q^b)=1$, when $s^b$ includes $s^a$, like the combination segment {\color[HTML]{ffd166}{\helvetica C}} including the past {\color[HTML]{ef476f}{\helvetica P}}.
\item $0<P(Q^a|Q^b)<1$, when $Q^a$ is possible but not necessary given by $Q^b$, like {\color[HTML]{ef476f}{\helvetica P}} and {\color[HTML]{118aff} \helvetica{F}} in Fig.~\ref{fig:latent_space}.
\item $0<P(Q^b|Q^a)<1$, defined in a similar manner.
\item $P(Q^a|Q^b)=0$, for unrelated segments, like {\color[HTML]{ffd166} {\helvetica C}} and {\color[HTML]{06d6a0} {\helvetica O}}.
\end{enumerate}
We treat the first three cases as positives, and the last case as a negative. Because pairs belonging to the first case have a stricter correspondence than those belonging to the second and third cases, we sample them more often during training. Note that under this interpretation, a past {\color[HTML]{ef476f}{\helvetica P}} and a future {\color[HTML]{118aff} \helvetica{F}} from the same trajectory do not have a strong correspondence (first case), but a softer one (second and third cases): one does not fully define the other. This approach results in probability values that we either maximize (positives) or minimize (negatives), so we define $D(A,B)=1-P(A|B)$.

The previous approaches require a way of computing either a distance between the distributions $Q$, or a conditional probability between them. In the next section, we show two families of distributions for which these can be defined. 

\subsection{Trajectory Segments as Distributions} 

In order to obtain $Q$, the encoder $\Theta$ predicts the parameters of a distribution family. Conditions for the distribution families are: 1) we can sample from it in a differentiable way, 2) we can parameterize it, 3) we can compute, in closed form, an intersection that returns a distribution from the same family, and 4) we can compute either a similarity function or a conditional probability, or both (see Sec.\ref{sec:approaches}). Next, we introduce two distribution families that meet the previous criteria.

\paragraph{Normal distributions} We use uncorrelated multivariate normal distributions, and parameterize them with a mean $\bm{\mu}$ and a standard deviation $\bm{\sigma}$. We compute the intersection as the product of two normal distributions, which remains normal when the dimensions are uncorrelated (see Appendix~\ref{apx:distributions}). We use the symmetrized Kullback-Leibler (KL) divergence between distributions as a distance function. This distance is not a proper metric; alternatives are discussed in Appendix~\ref{apx:distributions}. Normal distributions assume an underlying Euclidean distance metric $d$ between trajectories $z$.

\paragraph{Box embeddings}  Box embeddings  \cite{vilnis2018probabilistic} represent objects with high-dimensional products-of-intervals (or boxes), parameterized by their two extreme vertices $z^{\wedge}$ and $z^{\vee}$. The intersection between box embeddings is well defined and results in another box embedding. This makes them a natural choice to represent conditional probabilities, which can be computed as $P(A|B)=\mathrm{Vol}(A\cap B)/\mathrm{Vol}(B)$, where $\mathrm{Vol}(A)=\prod_{i}^N \max(z_i^{\vee}-z_i^{\wedge}, 0)$ is the volume of the box, and $\cap$ represents the intersection operation. These operations are straightforward to compute.
Boxes are not actual distributions, as they need not integrate to one. However, they are easily normalized by dividing by their volume, and therefore they can be treated as distributions for all the practical purposes required in our framework (\textit{i.e.} sampling, where we approximate the boxes with a uniform distribution). Symmetric distance functions can also be defined on box embeddings; we define a few in Appendix~\ref{sec:apx_boxes}.

In both cases, we use the reparameterization trick \cite{kingma2013auto} in order to sample from the distributions while keeping gradient information. We found the best-performing option was using box embeddings under the conditional scenario; the values reported in Section~\ref{sec:quantitative} use this setting.

\subsection{Inference}

Once trained, our decoder $\Phi$ is able to decode a trajectory at any continuous time $t$, including times that were not part of the input. For example, our framework can decode a future segment given an input past segment, by sampling from its representation, and evaluating that sample at some future times. This future segment will not necessarily be equal to the ground truth future segment (in case it exists), because a single past can have multiple futures. 

Overall, our framework is capable of doing 1) future and past prediction, by decoding a segment at times outside of its range; 2) continuous reconstruction given a discrete input, by decoding at any continuous time $t$; 3) interpolation between two segments, by decoding trajectories in their latent-space intersection; and 4) modifying existing trajectories, by manipulating the latent space. All the previous tasks are possible without explicitly training to do any of them. We show examples in Sec.~\ref{sec:experiments}. 

\section{Experiments}
\label{sec:experiments}

\begin{table}
\caption{\textbf{Prediction results}. We report the mean squared error (the lower the better) across keypoints, after normalizing each trajectory to be contained in a region of size $100\times100$. F, P and I stand for ``future'', ``past'' and ``interpolation'', respectively. Values are obtained over 10 runs with different test-time random seeds (changes include sampled segments and sampled $z$). An extended table with standard deviations is in Appendix~\ref{sec:additional}.}
\label{tab:results}
\begin{subtable}{1\linewidth}
\centering
\tabcolsep=0.11cm
  \caption{Long sequences}
  \label{tab:long}

    \begin{tabular}{l c ccc c  ccc c ccc}

    \toprule
    && \multicolumn{3}{c}{\textbf{FineGym}} && \multicolumn{3}{c}{\textbf{Diving48}} && \multicolumn{3}{c}{\textbf{FisV}} \\
    \cmidrule(r){3-5}             \cmidrule(r){7-9}              \cmidrule(r){11-13}
    &&  F      & P       & I       && F      & P       & I       && F      & P       & I       \\
        \midrule

\textbf{VRNN} \cite{vrnn}       &       & \s{ 15.85 }& \s{ 15.93 }& \s{ 16.10 }&        & \s{ 23.51 }& \s{ 27.97 }& \s{ 25.66 }&        & \s{ 14.95 }& \s{ 15.03 }& \s{ 15.08 }\\
\textbf{Trajectron++ uni.} \cite{salzmann2020trajectron++}       &       & \s{ \ 9.54 }& \s{ \ 9.98 }& \s{ \ 9.73 }&     & \s{ 11.67 }& \s{ 16.52 }& \s{ 11.98 }&        & \s{ 11.42 }& \s{ 11.85 }& \s{ 11.68 }\\
\textbf{Trajectron++} \cite{salzmann2020trajectron++}   &       & \s{ \ 9.72 }& \s{ 10.01 }& \s{ \ 9.89 }&      & \s{ 11.59 }& \s{ 16.23 }& \s{ 12.68 }&        & \s{ 11.41 }& \s{ 11.71 }& \s{ 11.63 }\\
\textbf{TrajRep (ours, ablation)}     &       & \s{ \ 8.82 }& \s{ \ 9.07 }& \s{ \ 7.57 }&     & \s{ 10.00 }& \s{\textbf{ 11.74 }}& \s{ 10.06 }&        & \s{ 10.62 }& \s{ 11.27 }& \s{ \ 9.70 }\\
\textbf{\quad + re-encoding (ours)} &       & \s{\textbf{ \ 8.50 }}& \s{\textbf{ \ 8.83 }}& \s{\textbf{ \ 7.11 }}&     & \s{\textbf{ \ 9.81 }}& \s{ 12.00 }& \s{\textbf{ \ 9.58 }}&      & \s{\textbf{ 10.32 }}& \s{\textbf{ 10.77 }}& \s{\textbf{ \ 9.22 }}\\
    \bottomrule
  \end{tabular}

\vspace{0.1cm}
    \caption{Short sequences}
    \label{tab:short}

    \begin{tabular}{l c ccc c  ccc c ccc}

    \toprule
    && \multicolumn{3}{c}{\textbf{FineGym}} && \multicolumn{3}{c}{\textbf{Diving48}} && \multicolumn{3}{c}{\textbf{FisV}} \\
    \cmidrule(r){3-5}             \cmidrule(r){7-9}              \cmidrule(r){11-13}
    &&  F      & P       & I       && F      & P       & I       && F      & P       & I       \\
        \midrule

\textbf{VRNN} \cite{vrnn}       &       & \s{ 12.77 }& \s{ 13.20 }& \s{ 13.40 }&        & \s{ 18.36 }& \s{ 20.14 }& \s{ 19.86 }&        & \s{ 13.26 }& \s{ 13.44 }& \s{ 13.45 }\\
\textbf{Trajectron++ uni.} \cite{salzmann2020trajectron++}       &       & \s{ \ 7.80 }& \s{ \ 8.28 }& \s{ \ 7.48 }&     & \s{ \ 9.05 }& \s{ 10.36 }& \s{ \ 8.29 }&      & \s{ \ 9.23 }& \s{ \ 9.68 }& \s{ \ 8.86 }\\
\textbf{Trajectron++} \cite{salzmann2020trajectron++}   &       & \s{ \ 7.26 }& \s{ \ 7.93 }& \s{ \ 6.94 }&     & \s{ \ 8.74 }& \s{ 11.35 }& \s{ \ 8.31 }&      & \s{ \ 8.70 }& \s{ \ 9.28 }& \s{ \ 8.28 }\\
\textbf{TrajRep (ours, ablation)}     &       & \s{ \ 6.49 }& \s{ \ 6.59 }& \s{ \ 5.15 }&     & \s{ \ 6.94 }& \s{ \ 6.99 }& \s{\textbf{ \ 5.00 }}&     & \s{ \ 7.83 }& \s{ \ 8.17 }& \s{ \ 6.01 }\\
\textbf{\quad + re-encoding (ours)} &       & \s{\textbf{ \ 6.20 }}& \s{\textbf{ \ 6.36 }}& \s{\textbf{ \ 4.88 }}&     & \s{ \textbf{\ 6.76 }}& \s{\textbf{ \ 6.85 }}& \s{ \ 5.04 }&     & \s{\textbf{ \ 7.54 }}& \s{\textbf{ \ 7.78 }}& \s{\textbf{ \ 5.88 }}\\
    \bottomrule
  \end{tabular}

\end{subtable}%

\end{table}

\subsection{Datasets}
\label{sec:datasets}

For our experiments, we selected data adhering to the following criteria. 
First, there has to be uncertainty in the trajectory when given just a segment (for instance, the future is not fully specified given the past). 
Second, the prediction should not require external contextual information. Context can be seamlessly added to our architecture, but it involves additional task-specific engineering decisions, and we want our evaluation to be orthogonal to them. Similarly, we avoid trajectories that require highly-engineered point-level distances $\delta$.
Finally, we prefer our trajectories to be obtained from real-world data. 
For all the previous reasons, we implement our framework on \emph{human movement datasets}. 

Specifically, we extract keypoints from human action datasets using OpenPose \cite{openpose}. For every video, we keep the most salient human trajectories. This results in sequences of dimension $[L, 25, 2]$, where $L$ is the number of frames in the trajectory, 25 is the number of joints in a human skeleton extracted by OpenPose, and 2 corresponds to the number of spatial coordinates for every joint. We refer to the whole skeleton at every time-step ---the combination of all joints, resulting in a $K=50$-dimensional vector--- as a \emph{point}. As a distance function $\delta$ between points (\textit{i.e.} skeletons) we use $l^2$-norm distance per-joint, and average across all visible joints. We extract human movement trajectories from the FineGym \cite{finegym}, Diving48 \cite{diving48} and FisV \cite{fisv} datasets, which correspond to gymnastics, diving and figure skating, respectively.

For each of the datasets, we experiment with short sequences (up to 10 time-steps, or slightly over one second) and long ones (up to 30 time-steps, representing slightly under four seconds of the trajectory), and report results for both. We provide more details on the dataset creation in Appendix~\ref{apx:datasets}.

\begin{figure}

\begin{subfigure}{1\linewidth}
\centering
\vspace{-1cm}
  \includegraphics[width=\columnwidth]{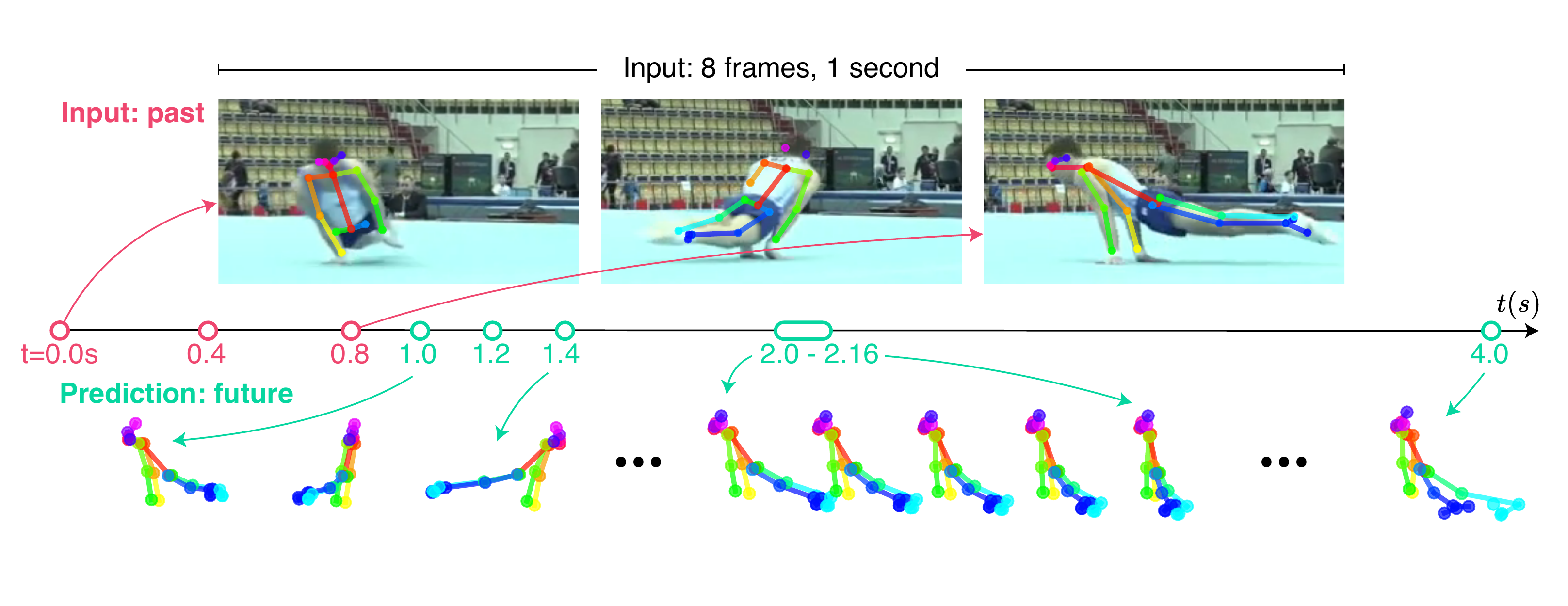}
      \vspace{-0.9cm}
  \caption{\textbf{Future prediction}. We show an example of a future prediction, where the input are eight irregularly sampled frames during one second (we only show three of them), and we predict up to three seconds into the future.}
    \vspace{-0.8cm}
  \label{fig:future_pred}
  \vspace{1cm}
    \includegraphics[width=\columnwidth]{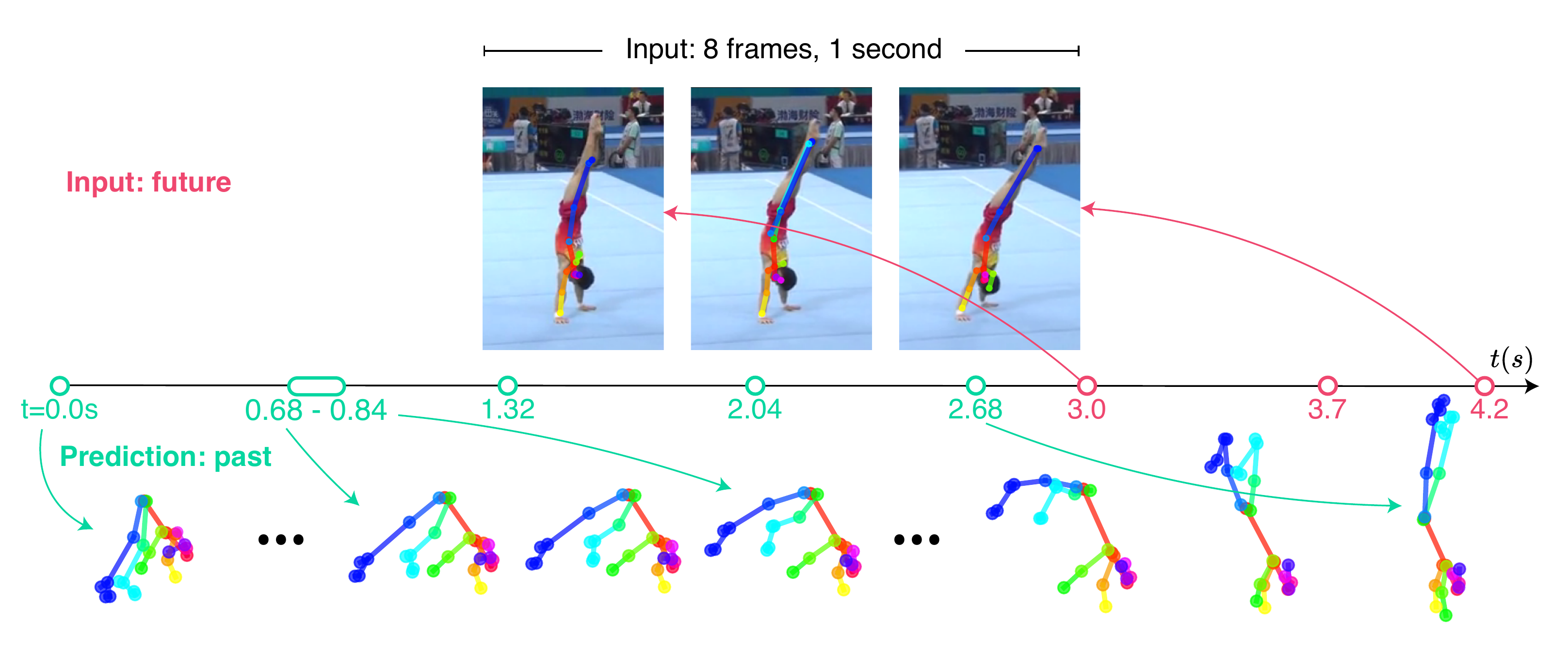}
        \vspace{-0.7cm}
  \caption{\textbf{Past prediction}. We show an example of a past prediction, where the input are eight irregularly sampled frames during one second (we only show three of them), and we predict up to three seconds into the past.}
  \label{fig:past_pred}
  \vspace{1cm}
    \includegraphics[width=\columnwidth]{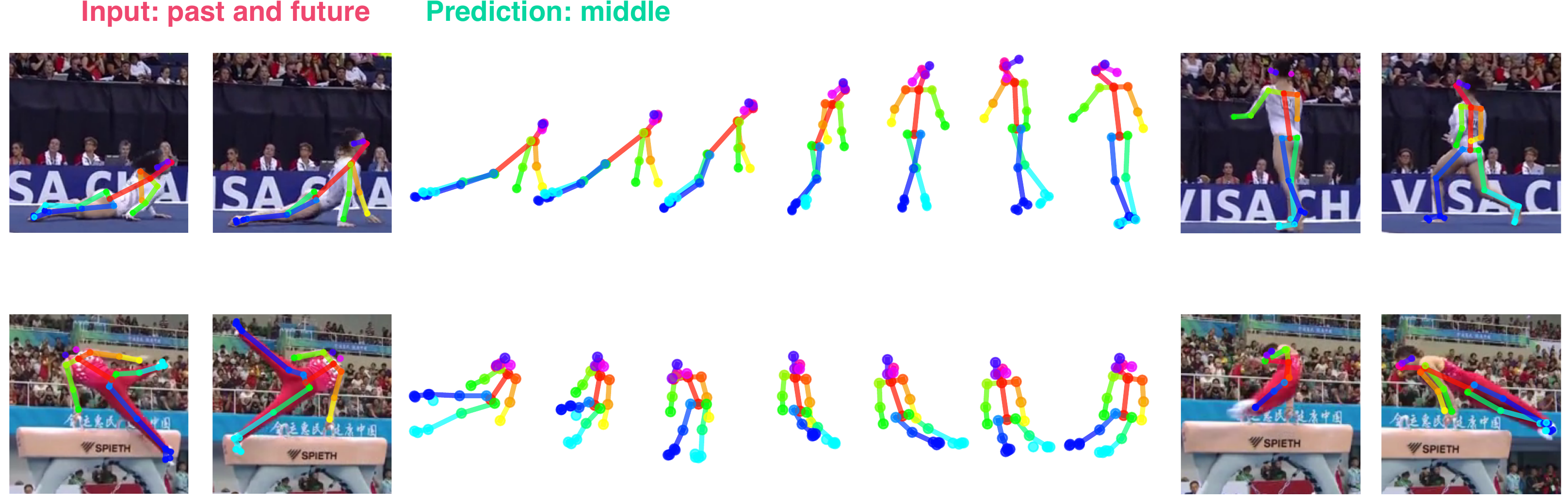}
  \caption{\textbf{Interpolation}. We provide the model with skeleton keypoints coming from two separate segments, and sample points at times in between these two segments, from the intersection of the two segments in latent space. The model produces sensible interpolations that are not simply a linear interpolation at the joint level: in the first example, the gymnast first stands, then turns; in the second example, the gymnast swings right and left, just in time to end up meeting the future segment at the right position.}
  \label{fig:interpolation}

\end{subfigure}
\vspace{0.2cm}
\caption{\textbf{Predictions on gymnastics data (FineGym)}. We show examples of past, future, and interpolation predictions. The input to our model are (irregularly in time) sampled keypoints obtained from human movement datasets, and the outputs are predictions of the trajectories at different continuous times (past, future, or in between the inputs). The only input to the model are the keypoints, not the images. Results show our model's capabilities for modeling trajectories well outside of the input's temporal range, for dealing with spatial and temporal occlusions, and for doing so at a large temporal resolution. See Section~\ref{sec:qualitative} for a deeper analysis.}
\label{fig:predictions}
\end{figure}

\subsection{Quantitative Experiments}
\label{sec:quantitative}

\paragraph{Baselines and ablations} As baselines, we select trajectory-modeling methods that are capable of encoding uncertainty about the future. \textbf{Variational RNNs} \cite{vrnn} extend recurrent neural networks (RNNs) to the non-deterministic case, by modeling every step with a variational auto-encoder (VAE) \cite{kingma2013auto}. \textbf{Trajectron++} \cite{salzmann2020trajectron++} is a state-of-the-art trajectory-modeling framework which also builds on top of RNNs and (conditional) VAEs \cite{kingma2014semi}. Uncertainty is modeled as a Gaussian mixture model (GMM). We adapt Trajectron++ to our data, making the encoding and decoding as similar to our setting as possible (for fairness), while keeping the core of the framework intact. We train two Trajectron++ versions, one with uniformly-sampled inputs and outputs (``Trajectron++ uni.''), and a second one with non-uniform sampling, following the setup in our models (``Trajectron++''). We also ablate our model, and report results with and without training with re-encoded segments.

\paragraph{Tasks and metric} We evaluate our framework on three different tasks: future prediction, past prediction, and interpolation between two segments. Future prediction consists in predicting points from a future segment given a past segment. Past prediction is defined symmetrically. In the interpolation task, we input two \emph{separated} segments (past and future) from a trajectory, and predict the segment in between them. In our model, we do so by decoding from the latent-space intersection of the two input segments. Baselines (which are autoregressive) are not capable of doing this combination, so we only use the past segment as input. Baselines are also not capable of directly performing past prediction, so we predict the future of the reversed trajectory instead.
As a metric, we use the average of the $l^2$-norm distance across joints, which is used by all methods during training, and report the best out of $M=10$ samples, to account for multiple modes and uncertainty in the prediction. Note that our model has never been explicitly trained to perform any of the previous tasks.

We show results in Tables~\ref{tab:long} and \ref{tab:short}, for long and short trajectories respectively. Our model outperforms baselines in all the tasks, which proves its value and flexibility. We also show how creating more interesting negatives with the re-encoding of decoded trajectories results in more accurate prediction results. However, our method performs well even without re-encoding.

\subsection{Qualitative Experiments}
\label{sec:qualitative}

\begin{figure}
\centering
\begin{subfigure}{0.8\linewidth}
\vspace{-0.5cm}
  \includegraphics[width=\columnwidth]{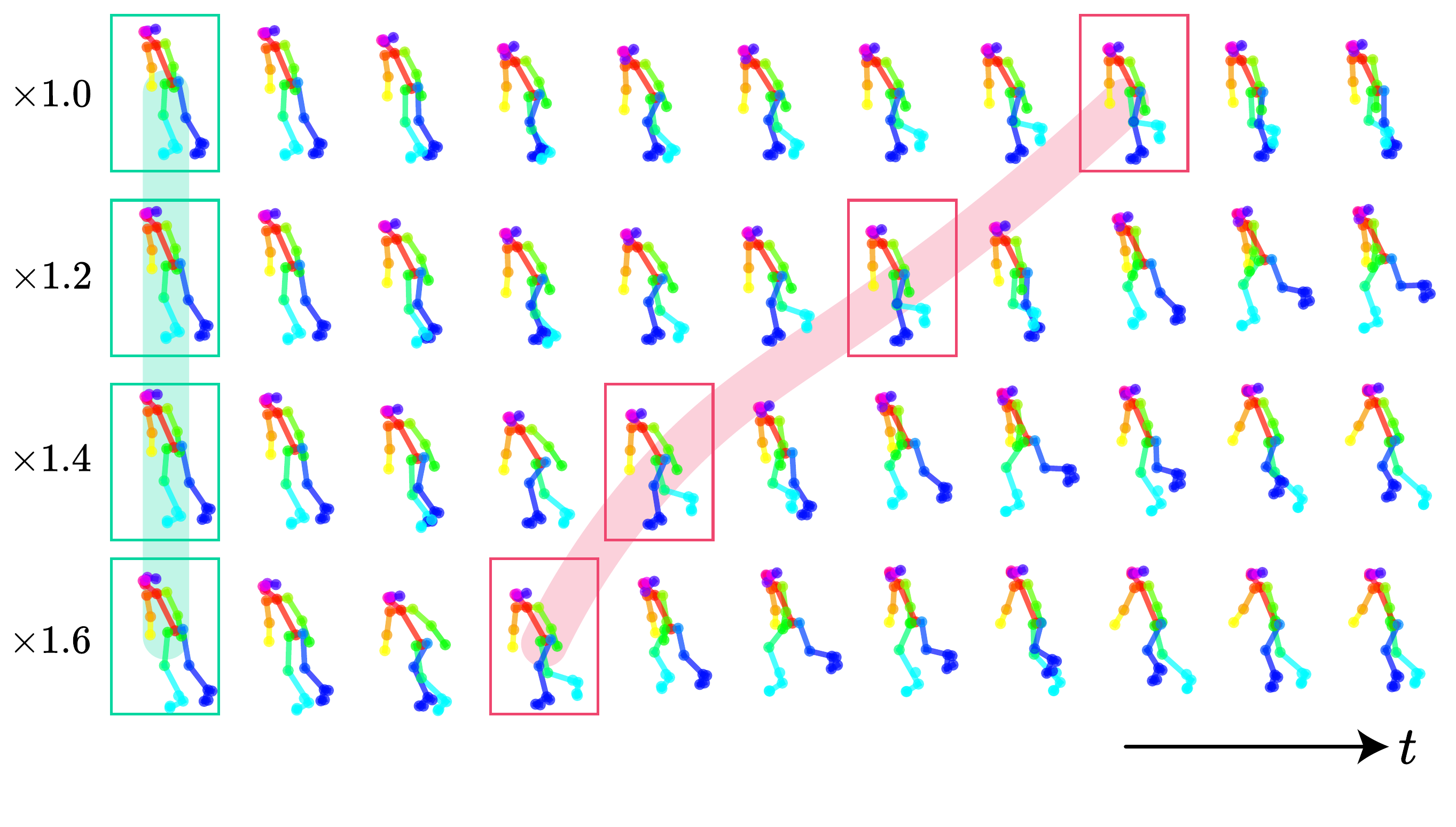}
  \vspace{-1cm}
  \caption{Speed change.}
  \label{fig:speed}
    \vspace{0.5cm}
  \centering
  \includegraphics[width=\columnwidth]{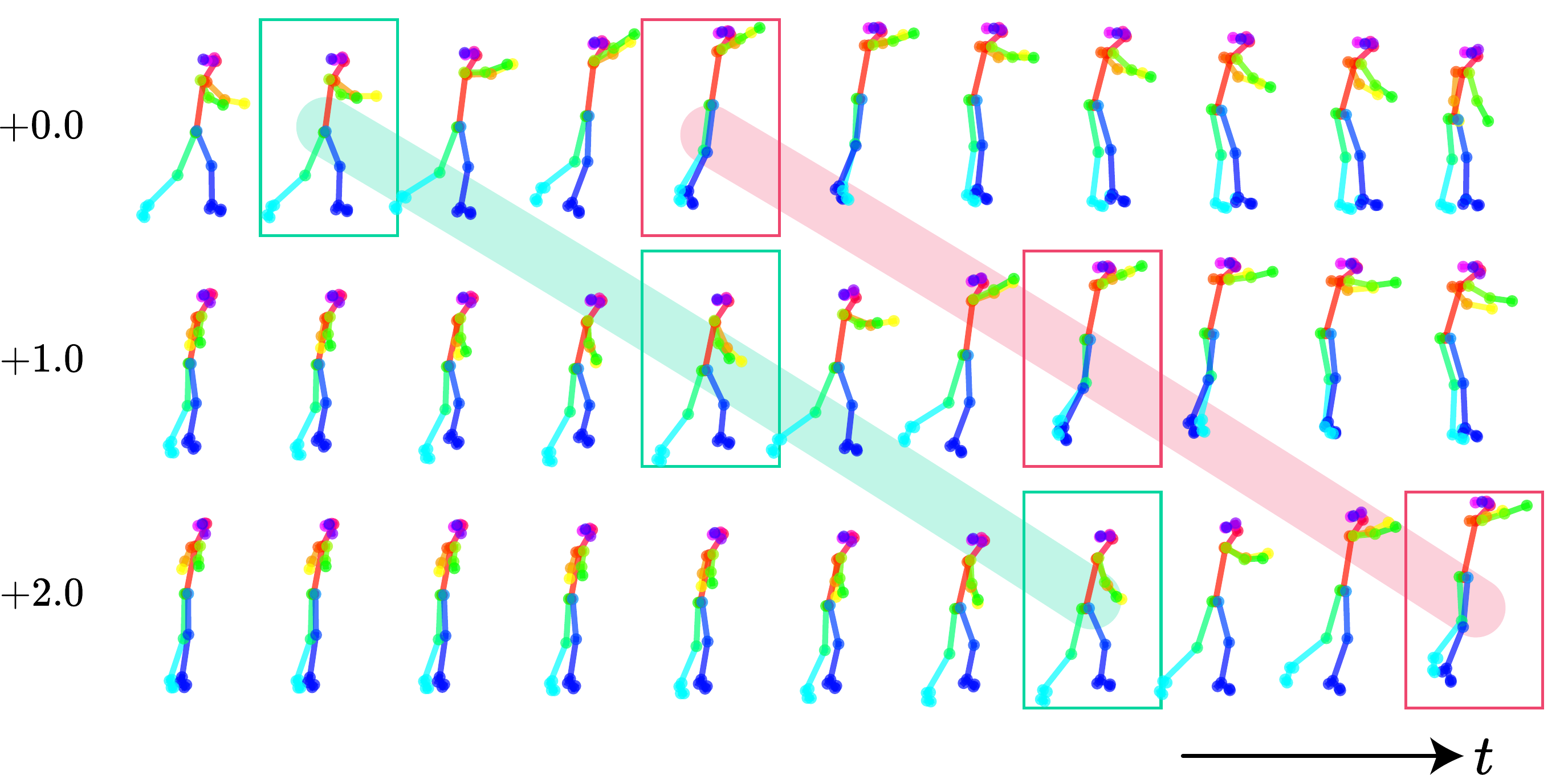}
  \caption{Temporal offset.}
  \label{fig:offset}
\end{subfigure}
\caption{\textbf{Temporal editing}. We decode, for the same times $t$, different trajectories $z$ in the latent space. These trajectories have been obtained by encoding the segment in the top row, and moving in small increments in the latent space along directions that represent speed (a) or temporal offset (b). We highlight in green and pink some correspondences between different decoded trajectories, to emphasize that the spatial trajectories are the same but with variations in some time-related attribute, such as speed or temporal offset.}
\label{fig:editing}
\end{figure}

\begin{figure}
  \centering
  \includegraphics[width=0.9\columnwidth]{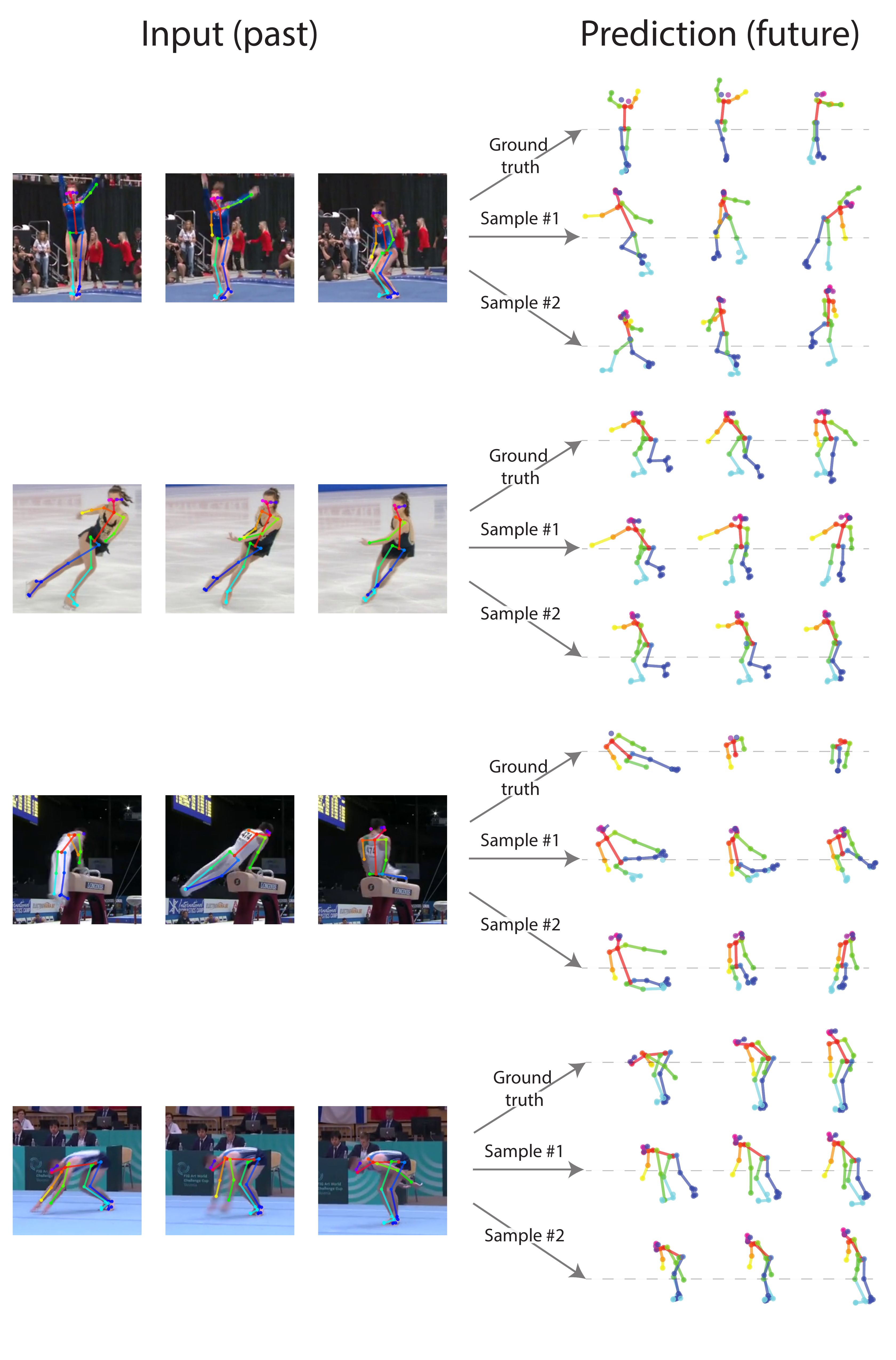}
  \caption{\textbf{Multiple futures}. Given a few input frames, our model is capable of predicting the future. It does so by modeling a distribution over the possible trajectories: by sampling from this distribution, we can obtain different plausible futures given the input (past) segment. In this figure we show, for specific inputs, the ground truth future, as well as two different futures sampled from the input segment distribution, which have been sampled randomly. This figure shows that our model is indeed capturing the multi-modal nature of the trajectories under uncertainty.}
  \label{fig:multimodes}
    \vspace{-0.2cm}
\end{figure}

We show examples of our model's inputs and outputs in Fig.~\ref{fig:predictions}. Specifically, we show future, past and interpolation predictions. The results reflect that the model learns to predict sequences up to four times longer than the input. They also show the large temporal resolution of our model: the model predictions evolve smoothly and sensibly for time-steps separated by a few hundredths of a second (\cref{fig:future_pred,fig:past_pred}). When not all joints are present in the input (first frame in Fig.~\ref{fig:future_pred}), our model is still capable of reconstructing the full spatial extent of the position. Finally, note how the model can take as inputs irregularly sampled time-steps (Fig.~\ref{fig:past_pred}), which makes it adaptable to temporal occlusions. Baselines are only capable of predicting the future, and are restricted to predicting uniform time-steps.

\paragraph{Temporal editing} The segments are directly tied to the temporal span they represent. For example, two segments with the exact same coordinates and evolution across time, but starting at different times will result in different (albeit similar) representations. The speed of a movement and the time in the trajectory where the movement is done are important attributes of that movement, and the representation should not be invariant to them: they belong to different trajectories. However, because these trajectories are very similar, our model learns to represent them close in the latent space. In Fig.~\ref{fig:editing}, we show that the model encodes different temporal variations. Moving along specific directions in the latent space results in progressively faster trajectories (\cref{fig:speed}), or in trajectories with an increasing temporal offset with respect to the original one (\cref{fig:offset}).
See Appendix~\ref{sec:additional} for details.

\paragraph{Representing multiple futures} A crucial aspect of our formulation is the assumption that the future is uncertain, and that our model has to be capable of modeling the different modes of the trajectory distribution. In Fig.~\ref{fig:multimodes} we show examples of multiple predicted futures given a single past segment, proving that our model captures the multi-modal nature of the trajectories under uncertainty.


\section{Related work}
\label{sec:related_work}

\paragraph{Modeling trajectories} Spatial trajectories are usually modeled in the literature in an autoregressive (AR) fashion \cite{song2017end,martinez2017human,kratzer2020prediction,alahi2016social,sun2019stochastic,ivanovic2019trajectron,salzmann2020trajectron++,li2019propagation,yan2018spatial,tang2019multiple,kosiorek2018sequential,hsieh2018learning}, where trajectories are defined conditioned on previous time-steps. Despite their success, AR models are incapable of dealing with some of the challenges stated in \cref{sec:introduction}, most notably they do not represent time as a continuous variable, they cannot model the full extent of a trajectory (simultaneously both past and future), and no learned trajectory-level metric can be obtained from them.
Some of them model the uncertainty in the prediction \cite{tang2019multiple,kosiorek2018sequential,hsieh2018learning,ivanovic2019trajectron,salzmann2020trajectron++,vrnn,sun2019stochastic}, and we use two representative ones \cite{vrnn,salzmann2020trajectron++} as our baselines.  
A different line of work is focused on representing segments of trajectories (not just points) as points in a latent space \cite{yao2017trajectory,zhang2020trajectory,yao2019computing,yao2020linear,li2018deep,zeng2020dsdnet,butepage2017deep,liu2022cstrm}. However, they are not capable of modeling the full extent of a trajectory outside the limits of the considered segment. Additionally, the segment-level metric is either unstructured \cite{yao2017trajectory}, or is explicitly given \cite{zhang2020trajectory,yuan2017review,zeng2020dsdnet}.

\paragraph{Continuous time} Modeling time as a continuous signal has gained traction recently in fields such as graphics \cite{xian2021space,vanhoorick2022revealing,pumarola2021d,sitzmann2019siren} or physics modeling \cite{michaelbeyond2022,chamberlain2021grand}, because it accurately represents the underlying (continuous) world being modeled. In the graphics neural implicit functions literature, time is used to condition the prediction of the network. We adopt the same approach in our decoder. We encode the set of continuous times in a segment by using a Transformer network \cite{transformers}, which by construction is permutation invariant, but allows temporal embeddings to be concatenated with the input, both discrete \cite{gberta_2021_ICML,arnab2021vivit} and continuous \cite{vanhoorick2022revealing,fourier_encodings}.


\paragraph{Self-supervised representation learning} Finding self-supervised representations for temporal data has been the subject of a large amount of work in domains such as trajectories \cite{liu2022cstrm}, video \cite{pan2021videomoco,han2019video,qian2021spatiotemporal}, or audio processing \cite{jansen2018unsupervised,saeed2021contrastive,fonseca2021unsupervised}. Most methods, however, represent segments as simple points in a Euclidean space. Structured representations for temporal data \cite{suris2021hyperfuture,park2022probabilistic} allow the latent space to follow certain inductive biases, like our framework's idea that segments compose trajectories. We model segments as either normal distributions or box embeddings \cite{vilnis2018probabilistic}. The latter have been used to represent hierarchical relationships in domans such as text \cite{patel2020representing,onoe2021modeling}, knowledge bases \cite{abboud2020boxe}, or images \cite{rau2020predicting}. We use them to represent temporal information. In recent work, Park et al. \cite{park2022probabilistic} also model segments using normal distributions, where trajectories are weighted sums of the segment representations.

\begin{ack}
We thank Arjun Mani and Mia Chiquier for helpful feedback. This research is based on work partially supported by the NSF NRI Award \#2132519 and the DARPA MCS program. DS is supported by the Microsoft PhD Fellowship.
\end{ack}

{
    \clearpage
    \small
    \bibliographystyle{ieee_fullname}
    \bibliography{main}
}

\clearpage
\appendix

\section{Limitations and Societal Impact}
\label{sec:discussion}

\paragraph{Limitations} Our approach assumes all the information about a trajectory can be represented by a single latent-space embedding. While this holds true in our experiments, it is unclear how well it can scale to more complex data (like pixel-space videos) or longer trajectories (of the order of minutes). See Subramani et al. \cite{subramani2019can} for a related discussion. Another limitation we noticed from our model is that, while it is in principle capable of generating diverse trajectories from given segments, it tends to predict ``average'' trajectories when asked to predict far into the future (or past). Baselines have a similar behavior. A generative approach to the decoding, which would naturally fit into our framework, would probably help generating diverse trajectories far into the future; we leave it to future work. Finally, we mention in Section~\ref{sec:datasets} that, by construction, contextual information can easily be added to the model, but we did not try it. We leave this exploration to future work. 

\paragraph{Negative societal impact} Our approach is data-driven, and thus it will replicate the biases seen in the data. Trajectory data is not as sensitive as image data on its own. However, trajectory-modeling methods can be used in sensible applications such as tracking of people for surveillance purposes.

\section{Dataset Details}
\label{apx:datasets}

\begin{table}
    \caption{\textbf{Dataset details.} }
    
    \centering
  \label{tab:dataset}

    \begin{tabular}{lcccccc}

    \toprule
    & \multicolumn{2}{c}{\textbf{FineGym}} & \multicolumn{2}{c}{\textbf{Diving48}} & \multicolumn{2}{c}{\textbf{FisV}} \\
    \cmidrule(r){2-3}             \cmidrule(r){4-5}              \cmidrule(r){6-7}
    & Long & Short & Long & Short & Long & Short \\
    \midrule
    Training trajectories &    404k & 665k & 787 & 27k & 231k & 276k \\
    Validation trajectories &  50k & 83k & 98 & 3k & 29k & 35k\\
    Test trajectories & 50k & 83k & 98 & 3k & 29k & 35k\\
    Total trajectories &  505k & 832k & 984 & 33647 & 289k & 345k\\
    Total videos &  \multicolumn{2}{c}{13k} & \multicolumn{2}{c}{18k} & \multicolumn{2}{c}{500}\\
    \bottomrule
  \end{tabular}
\end{table}

For each one of the considered datasets, we process all the frames individually with OpenPose \cite{openpose} to extract human skeleton keypoints. OpenPose returns the keypoints of all the detected humans. We then post-process the extracted keypoints, in order to group the ones belonging to the same person over time. We use heuristics that include spatial closeness between two consecutive frames and similar size of the skeleton. Once the correspondences have been obtained, and we get a series of trajectories, we filter out those that are either too small, too short, or too noisy. The minimum length (in seconds) of the trajectories is set to be the maximum length of the combination segments being sampled, so that, from the point of view of the segments, the trajectories are not limited in time. 

In the short-segment experiments, the segment length is up to 30 frames, from which we randomly (read: non-uniformly) sample a third of them (this is, up to 10), for slightly over a second long segments. In the long-segment experiments, the segment length is up to 90 frames, from which we randomly sample a third of them (this is, up to 30), for slightly under four seconds long segments. Note that our model is not limited to predicting trajectories within this time-span, and it can extrapolate to larger time values.

We randomly divide the obtained trajectories into training, validation and test splits in a 80/10/10 proportion. We report the size of each dataset in Table~\ref{tab:dataset}. Because Diving48 is a small dataset, we fine-tune the FineGym-trained models for it, both in our models and in baselines. The reason there are very few long trajectories for Diving48 is that OpenPose extracts noisy keypoints due to out-of-distribution positions, which get filtered-out in our post-processing. The amount of videos reported for FineGym and Diving48 corresponds to the number of events; there is more than one event in each original video, but we pre-process them into separate event-level videos before extracting the keypoint trajectories. 

\section{Implementation Details}
\label{apx:details}

\paragraph{Architecture}
We implement the encoder network $\Theta$ using a Transformer Encoder \cite{transformers} with two layers and two heads, and hidden size 512. The latent space dimensionality $N$ is also 512. The input to the Transformer are the spatial positions at every sampled time-step, as well as a \texttt{[CLS]} token to represent the whole trajectory. We use the output of the \texttt{[CLS]} token to represent the distributions $Q$. We append a temporal embedding to each input, corresponding to the time each point was sampled. We use Fourier encodings \cite{fourier_encodings} to encode the continuous time value to its vector representation. We found using MLPs had similar results but slightly worse generalization to extrapolation to out-of-distribution time-steps.

The decoder network $\Phi$ is implemented using a ResNet with four blocks and hidden size 512. The temporal indices are also encoded using Fourier encodings, and they are concatenated to the $z$ value to be decoded, following prior work \cite{vanhoorick2022revealing}. We use the same architecture for the decoder network in the baselines. The capacity of the baselines is the same as our method's (\textasciitilde 10M parameters).

\paragraph{Inputs}
In order to avoid learning shortcuts by $\Theta$, the combination segment has the same temporal span as the combination of the past and future segments, but the specific sampled points are different (different times), so that the model cannot simply identify that the values are exactly the same and use that information to bring representations together. For every trajectory in the dataset, we randomly sample a start and end point for the segments to be used.

\paragraph{Losses}
We classify segment pairs as hard positives, which we deem very important, soft positives, soft negatives, and hard negatives, and sample hard positives and negatives more often. For example, the pair consisting of the past {\color[HTML]{ef476f}{\helvetica P}} and its reconstruction {\color[HTML]{aa00ff} {\helvetica {PP}}} is deemed important enough to be a hard positive, and samples from other random elements in the batch are considered soft negatives. We found in initial experiments that the selection of which pairs are hard or soft is only marginally important, as long as the overall intuition described in Section~\ref{sec:training} is followed. We report the specific choices in ~\cref{fig:matrices}. Note that in the conditional setting, where we are minimizing and maximizing probability values, optimizing a binary cross-entropy loss instead of a triplet loss may seem more adequate. However, we found in initial experiments that the triplet loss performed better, so we kept it. 

We found using $l^1$-norm for the distance function $\delta$ to perform similarly to the $l^2$-norm, so we kept using $l^2$-norm but $l^1$-norm is a viable alternative. This shows that our model is general and not dependant on the specific $\delta$.

In the triplet loss in \cref{eq:traj}, we set the margin $\alpha=1$.

When sampling trajectories to be decoded, we sample 3 trajectories for every distribution $Q$. These will be either positives or negatives among themselves, depending on the case (see \cref{fig:matrices}).

\paragraph{Optimization} The model parameters are optimized using AdamW \cite{adamw} with weight decay of $0.05$. For the learning rate, we use a cosine annealing strategy, with ranges $[1e-6, 1e-4]$, a period of 4k steps, and a warmup of 1k steps. Additionally, we clip the gradients by norm for values larger than $0.01$. We did not run a hyperparameter search on the previous parameters, as we found that the initial ones worked well, except for the gradient clipping, which we found necessary to stabilize the training of the Transformer. Finally, we trained the model using mixed precision.

We use PyTorch \cite{pytorch}, and the box embeddings library \cite{chheda2021box}. The models take two days to train on four RTX208-Ti GPUs.

Code, models, and data are provided. 

\subsection{All pair-to-pair negative/positive relationships}
\label{sec:all_pairs}

\begin{figure}
\centering
\begin{subfigure}{0.49\linewidth}
  \centering
  \includegraphics[width=1\columnwidth]{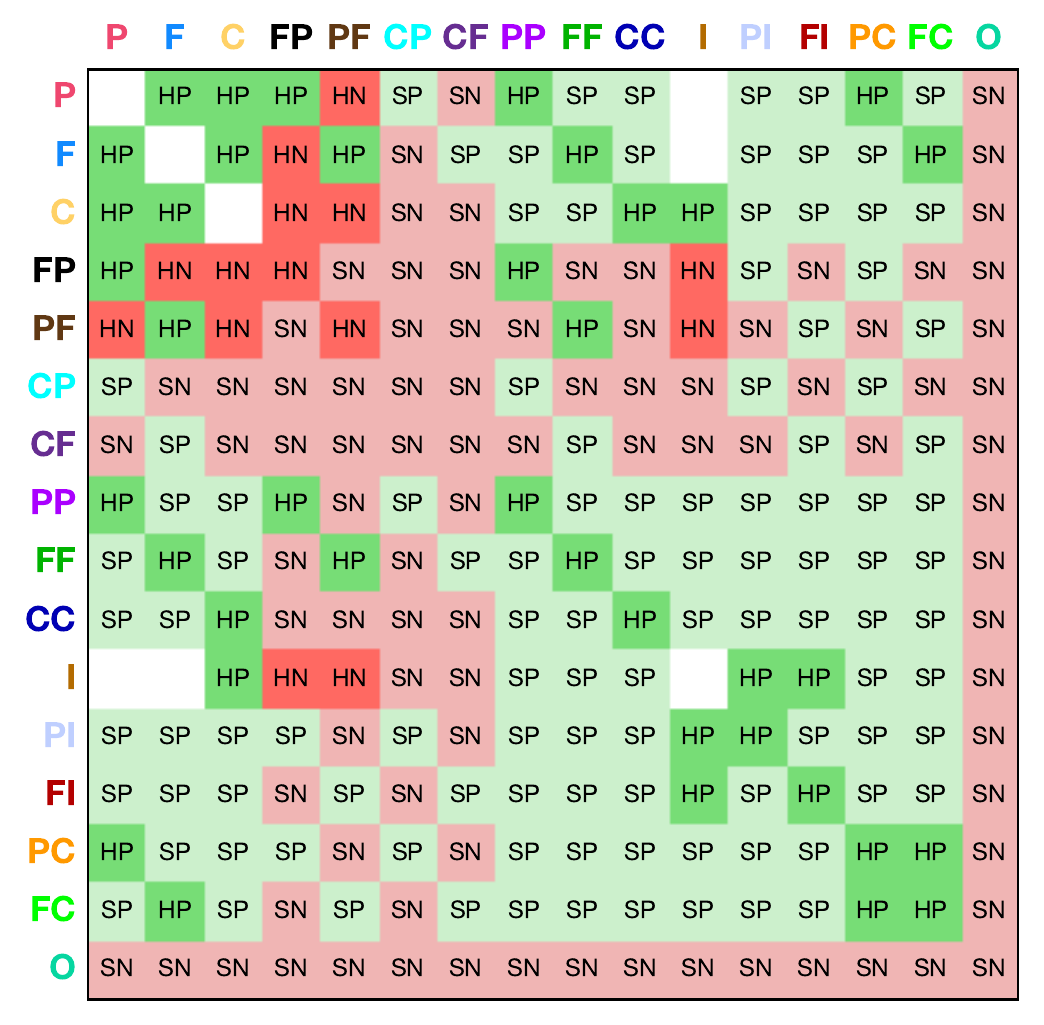}
  \caption{\textbf{Symmetric approach}}
\end{subfigure}
\begin{subfigure}{0.49\linewidth}
  \centering
  \includegraphics[width=1\columnwidth]{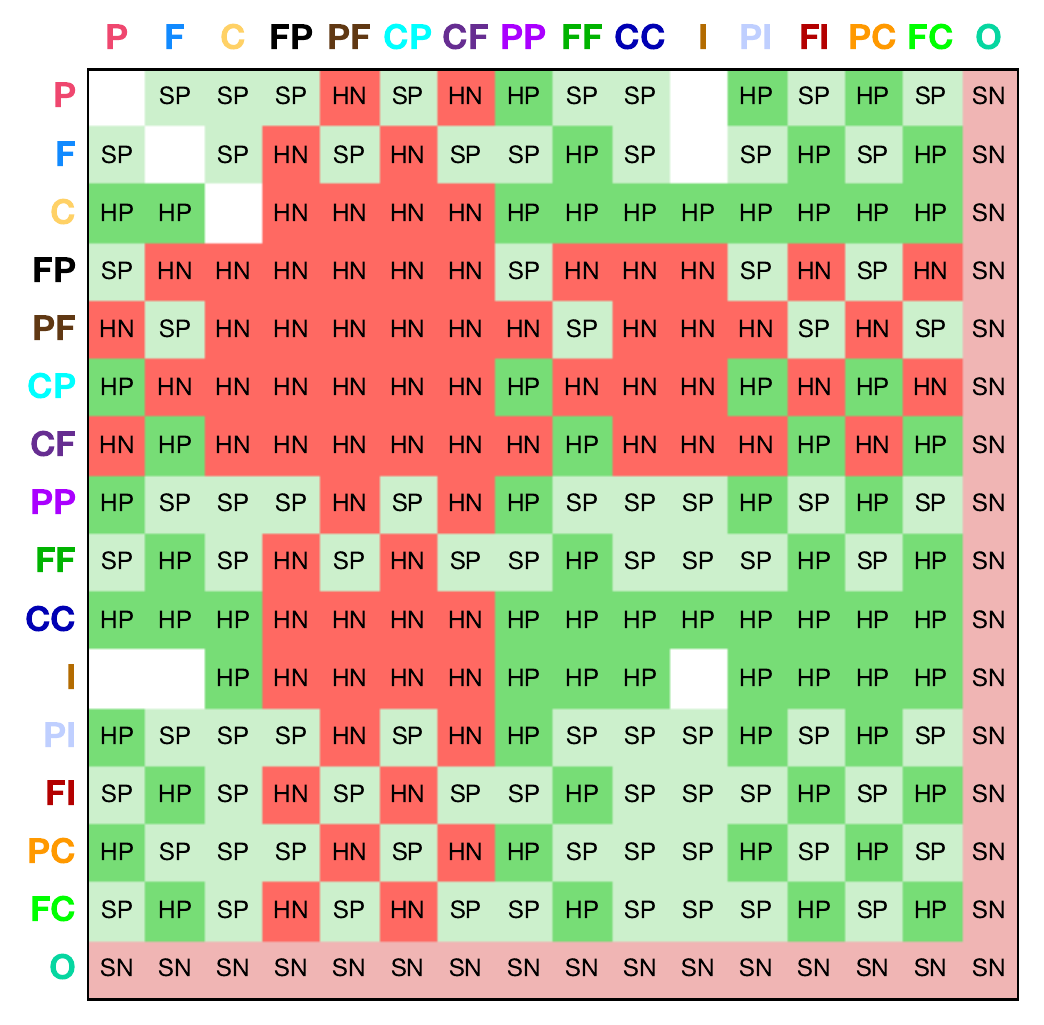}
  \caption{\textbf{Conditional approach}}
\end{subfigure}
\caption{\textbf{Positive and negative pairs}. HP, SP, SN, HN stand for ``hard positive'', ``soft positive'', ''soft negative'' and ``hard negative'', respectively. Note that the two matrices are the same if we consider hard and soft to be equal. The symmetric approach results in symmetric negatives and positives, while the conditional one results in an asymmetric one, because $P(A|B)\neq P(B|A)$ in general. However, as expected, it is symmetric if we consider hard and soft negatives to be equal. In that case, the two approaches result in the same matrices. Best seen in color.}
\label{fig:matrices}
\end{figure}

We follow the notation in Fig.~\ref{fig:latent_space}, and show pairwise negatives and positives in Fig.~\ref{fig:matrices}. Note that some segments have a negative or positive value defined with respect to themselves: this corresponds to the relationship between different samples. For example, given a past {\color[HTML]{ef476f}{\helvetica P}}, multiple futures {\color[HTML]{000000} {\helvetica {FP}}} can be defined, with a specific relationship between them (in this case, they will be hard negatives). Additionally, we add other options that are not shown in Fig.~\ref{fig:latent_space}. We list all of them here:
\begin{itemize}[topsep=0pt,itemsep=-1ex,partopsep=1ex,parsep=1ex,leftmargin=1cm,labelwidth={1em}]
\item [{\color[HTML]{ef476f}{\helvetica P}}] Past.
\item [{\color[HTML]{118aff} \helvetica{F}}] Future.
\item [{\color[HTML]{ffd166} {\helvetica C}}] Combination.
\item [{\color[HTML]{000000} {\helvetica {FP}}}] Future given past. Representation that has been obtained from encoding a segment that has been decoded from the past {\color[HTML]{ef476f}{\helvetica P}} representation at times $t$ obtained from the future {\color[HTML]{118aff} \helvetica{F}} segment.
\item [{\color[HTML]{603813} {\helvetica {PF}}}] Past given future.
\item [{\color[HTML]{00ffff} {\helvetica {CP}}}] Combination given past.
\item [{\color[HTML]{662d91} {\helvetica {CF}}}] Combination given future.
\item [{\color[HTML]{aa00ff} {\helvetica {PP}}}] Past given past. As in the previous cases, this corresponds to decoding the past  {\color[HTML]{ef476f}{\helvetica P}} representation into the times in the past segment, and then re-encoding the obtained segment.
\item [{\color[HTML]{00b300} {\helvetica {FF}}}] Future given future.
\item [{\color[HTML]{0000b3} {\helvetica {CC}}}] Combination given combination.
\item [{\color[HTML]{b36b00} {\helvetica {I}}}] Intersection. Distribution that results from intersecting the past {\color[HTML]{ef476f}{\helvetica P}} and future {\color[HTML]{118aff} \helvetica{F}} representations in the latent space.
\item [{\color[HTML]{bfcfff} {\helvetica {PI}}}] Past given intersection.
\item [{\color[HTML]{b30000} {\helvetica {FI}}}] Future given intersection.
\item [{\color[HTML]{ff9900} {\helvetica {PC}}}] Past given combination.
\item [{\color[HTML]{00ff00} {\helvetica {FC}}}] Future given combination.
\item [{\color[HTML]{06d6a0} {\helvetica O}}] Other (segment from a different trajectory in the batch).
\end{itemize}


In practice, not all positive and negative pairs are equally important. We differentiate between soft and hard positives and negatives, where hard ones are deemed more relevant and we give them a more important role in the optimization, by sampling them more often during training. The motivation behind distinguishing between hard and soft pairs is that we want to mostly rely on strong signals, and not add too much noise to the training. However, the distinction between hard and soft pairs is not as crucial and fundamental to our framework as the distinction between positives and negatives. Different criteria to select hard and soft pairs could be used, and the rules we used to differentiate between hard and soft positives and negatives are as follows. First, let us define the representations of segments that come directly from the data as ``first encodings'', to distinguish them from the re-encoded ones.

For the symmetric approach, where positives are defined as those segments that can belong to the same trajectory, \emph{hard} positives are the positive pairs where \textit{either} of these conditions is met:
\begin{itemize}[topsep=0pt,itemsep=-1ex,partopsep=1ex,parsep=1ex,leftmargin=1cm,labelwidth={1em}]
\item The pair consists of two first encodings. For example, ({\color[HTML]{ef476f}{\helvetica P}}, {\color[HTML]{118aff} \helvetica{F}}).
\item Pairs that encode \textit{exactly} the same segment, and thus the distribution should be exactly the same. For example, ({\color[HTML]{ef476f}{\helvetica P}}, {\color[HTML]{aa00ff} {\helvetica {PP}}}).
\item Additionally, we add four extra hard positives that do not meet either of the previous conditions, but are necessary so that some hard negatives can act as such. In a triplet loss, a (hard) negative requires a (hard) positive to be contrasted to. Some hard negatives like ({\color[HTML]{603813} {\helvetica {PF}}}, {\color[HTML]{603813} {\helvetica {PF}}}) would not have a hard positive associated to them if we only followed the two previous conditions ({\color[HTML]{603813} {\helvetica {PF}}} would not be hard positive with any other segment), and for this reason we explicitly create these four hard positives, which are ({\color[HTML]{ef476f}{\helvetica P}}, {\color[HTML]{000000} {\helvetica {FP}}}), ({\color[HTML]{118aff} \helvetica{F}}, {\color[HTML]{603813} {\helvetica {PF}}}), ({\color[HTML]{aa00ff} {\helvetica {PP}}}, {\color[HTML]{000000} {\helvetica {FP}}}) and ({\color[HTML]{00b300} {\helvetica {FF}}}, {\color[HTML]{603813} {\helvetica {PF}}}).
\end{itemize}

\emph{Hard negatives} in the symmetric case are the negative pairs where \textit{both} of these conditions are met:
\begin{itemize}[topsep=0pt,itemsep=-1ex,partopsep=1ex,parsep=1ex,leftmargin=1cm,labelwidth={1em}]
\item One element of the pair represents either the past or the future.
\item The other element is either a first encoding (\emph{e.g.} the pair ({\color[HTML]{603813} {\helvetica {PF}}}, {\color[HTML]{ef476f}{\helvetica P}})), or it represents the same segment as the first element of the pair, but a different sample (\emph{e.g.} the pair ({\color[HTML]{603813} {\helvetica {PF}}}, {\color[HTML]{603813} {\helvetica {PF}}})).
\end{itemize}

For the conditional approach, the distinction between hard and soft positives is more clear: hard positives are those where $P(A|B)=1$ (this is, the first case in the list, where given $B$ we can be certain of $A$), for example for $P({\color[HTML]{ef476f}{\helvetica P}}|{\color[HTML]{ffd166} {\helvetica C}})=1$. Other positives (second and third cases) are treated as soft. All negatives are treated as hard negatives except for the ones coming from different elements in the batch.

An ablations where we make all positives hard is included in Tab.~\ref{tab:ablations}.

\section{Distribution Families}
\label{apx:distributions}

\subsection{Normal Distributions}

Multivariate normal (also called Gaussian) distributions have the following probability density function (PDF):

\begin{equation}
    p(\mathbf{z};\boldsymbol{\mu},\boldsymbol{\Sigma})=\frac{1}{\sqrt{(2\pi)^N|\boldsymbol{\Sigma}|}}\exp\left[-\frac{1}{2}(\mathbf{z}-\boldsymbol{\mu})^T\boldsymbol{\Sigma}^{-1}(\mathbf{z}-\boldsymbol{\mu})\right],
\end{equation}
where $\mathbf{z}$ is a real $N$-dimensional column vector, $\mathbf{\mu}$ is the mean, and $\mathbf{\Sigma}$ is the covariance matrix.  
In the main paper all variables $z$ and $x$ are vectors, so we do not use their bold versions $\mathbf{z}$ and $\mathbf{x}$, as they cannot possibly be confused with scalars.

We assume an uncorrelated multivariate normal distribution, so the previous equation is simplified to:
\begin{equation}
    p(\mathbf{z};\boldsymbol{\mu},\boldsymbol{\sigma})=\prod_{i=1}^N\frac{1}{\sqrt{2\pi\sigma_i^2}}\exp\left[-\frac{(z_i-\mu_i)^2}{2\sigma_i^2}\right],
\end{equation}
where $\boldsymbol{\sigma}$ is a vector representing the individual-dimension standard deviations.

The product of two normal PDFs (\emph{not} the product of two normal random variables) results in a scaled normal distribution when the $N$ variables are uncorrelated. This is \emph{not} general for other multivariate normal distributions. See Bomiley \cite{bromiley2003products} for derivations in the two scenarios. The resulting PDF is the product of the individual dimension density functions, which follow the equation: 
\begin{equation}
\begin{split}
p(z) q(z)=\frac{S_{p q}}{\sqrt{2 \pi \sigma_{p q}^2}} &\exp \left[-\frac{\left(z-\mu_{p q}\right)^{2}}{2 \sigma_{p q}^{2}}\right], \\
\textrm{where}\quad\sigma_{p q}=\sqrt{\frac{\sigma_{p}^{2} \sigma_{q}^{2}}{\sigma_{p}^{2}+\sigma_{q}^{2}}} \quad &\textrm{and} \quad \mu_{p q}=\frac{\mu_{p} \sigma_{q}^{2}+\mu_{q} \sigma_{p}^{2}}{\sigma_{p}^{2}+\sigma_{q}^{2}}.
\end{split}
\end{equation}
In the previous equation, $p(z)$ and $q(z)$ are univariate normal PDFs with mean and variance $\mu_p, \sigma_q$ and $\mu_p, \sigma_q$, respectively, and the scaling factor $S_{p q}$ is itself a normal PDF on both $\mu_p$ and $\mu_q$ with standard deviation $\sqrt{\sigma_p^2+\sigma_q^2}$:
\begin{equation}
    S_{p q}=\frac{1}{\sqrt{2\pi(\sigma_p^2+\sigma_q^2)}}\exp\left[-\frac{(\mu_p-\mu_q)^2}{2(\sigma_p^2+\sigma_q^2)}\right].
\end{equation}
In order to compute the intersection of the two original normal PDFs, we simply ignore the scaling factor.

\paragraph{Kullback–Leibler divergence} For two distributions $P$ and $Q$ of a continuous random variable, the Kullback–Leibler (KL) divergence is defined to be the integral:
\begin{equation}
   D_{\mathrm{KL}}(P \| Q)=\int_{-\infty}^{\infty} p(z) \log \frac{p(z)}{q(z)}\ d z,
\end{equation}
where $p$ and $q$ denote the probability densities of $P$ and $Q$. The KL divergence measures how well a probability distribution $Q$ represented another distribution $P$. In the specific case of normal distributions, the previous equation is \cite{hershey2007approximating}:
\begin{equation}
\begin{split}
    D_{\mathrm{KL}}(P\|Q)&=\frac{1}{2}\left[\log \frac{\left|\boldsymbol{\Sigma}_{q}\right|}{\left|\boldsymbol{\Sigma}_{p}\right|}-N+\operatorname{Tr}\left[\boldsymbol{\Sigma}_{q}^{-1} \boldsymbol{\Sigma}_{p}\right]+\left(\boldsymbol{\mu}_{q}-\boldsymbol{\mu}_{p}\right)^{T} \boldsymbol{\Sigma}_{q}^{-1}\left(\boldsymbol{\mu}_{q}-\boldsymbol{\mu}_{p}\right)\right] \overset{\mathrm{uncorrelated}}{=}\\&=\sum_i^N\log{{\sigma_q}_i}-\log{{\sigma_p}_i}-\frac{1}{2}+\frac{{{\sigma_p}_i}^2+{\left({\mu_q}_i-{\mu_p}_i\right)^2}}{2{{\sigma_q}_i}^2}.
    \end{split}
\end{equation}

Note that $D_{\mathrm{KL}}(P\|Q)$ is technically not a distance metric, just a divergence. Also, $D_{KL}(p\|q)$ is asymmetric, so in the symmetric approach we use the symmetrized version $D_{\mathrm{KL}}(P,Q)=(D_{\mathrm{KL}}(P\|Q)+D_{\mathrm{KL}}(Q\|P))/2$.

Proper distance metrics between normal distributions can be defined. For example, the $p^{\textrm{th}}$ Wasserstein distance between two distributions $P$ and $Q$ is generally defined as:
\begin{equation}
    W_{p}(P, Q):=\left(\inf _{\gamma \in \Gamma(P, Q)} \int_{Z \times Z} d(z_p, z_q)^{p} \mathrm{~d} \gamma(z_p, z_q)\right)^{1 / p}
\end{equation}
where $\Gamma(P, Q)$ denotes the collection of all measures on $Z\times Z$ with marginals $P$ and $Q$ on the first and second factors, respectively. When the underlying distance metric $d$ is the $l^2$-norm distance function, the Wasserstein distance between normal distributions has the closed-form expression \cite{salmona2021gromov}: 

\begin{equation}
\begin{split}
    W_{2}^{2}(P,Q)&=\left\|\boldsymbol{\mu}_{q}-\boldsymbol{\mu}_{p}\right\|^{2}+\operatorname{Tr}\left[\boldsymbol{\Sigma}_{p}+\boldsymbol{\Sigma}_{q}-2\left(\boldsymbol{\Sigma}_{p}^{\frac{1}{2}} \boldsymbol{\Sigma}_{q} \Sigma_{p}^{\frac{1}{2}}\right)^{\frac{1}{2}}\right]\overset{\mathrm{uncorrelated}}{=}\\
    &=\sum_i^N({\mu_q}_i-{\mu_p}_i)^{2}+{\sigma_p}_i^2+{\sigma_q}_i^2-2{\sigma_p}_i{\sigma_q}_i.
    \end{split}
\end{equation}

Defining a distribution metric using an optimal transport approach has two main advantages. First, it results in an actual metric, as opposed to just a divergence or non-metric distance function. And second, it makes it very explicit what the underlying distance function $d$ is in the latent space. However, it is less clear than in the KL case that minimizing the $W_2$ distance between normal distributions will result in a large overlap between them, as opposed to just spatial proximity. Therefore, we use KL divergence instead.



\subsection{Box Embeddings}
\label{sec:apx_boxes}

Box embeddings are $N$-dimensional hyperrectangles that can represent relationships such as intersection and containment. Boxes are parameterized by their two extreme points $z^{\wedge}$ and $z^{\vee}$. They are designed to represent unary and joint probabilities of events (segments, in our case), where large boxes represent highly probable and general concepts.
The original paper \cite{vilnis2018probabilistic} defines boxes as step functions (rectangles), but posterior papers smooth the edges of the boxes so that all pairs of boxes have positive intersections. Specifically, Li et al. \cite{li2018smoothing} use Gaussian convolutions, and Dasgupta\&Boratko et al. \cite{dasgupta2020improving} improve on the previous paper using min and max Gumbel distributions. This results in better gradients during optimization, while keeping the intuition and parameters the same. We use the latter, which is conveniently implemented in an open source library \cite{chheda2021box}. 

We sample from a box by assuming the edges are hard instead of soft, and assuming a uniform distribution in the range $[z^{\wedge}, z^{\vee}]$. The volume of a box is computed as: 
\begin{equation}
    \mathrm{Vol}(A)=\prod_{i}^N \max(z_i^{\vee}-z_i^{\wedge}, 0)
\end{equation}
In practice, the $\max$ operation is replaced by soft versions. The intersection between two boxes $A$ and $B$ can be computed as $z_{\cap }^{\wedge}= \max(z_a^{\wedge}, z_b^{\wedge})$ and $z_{ \cap }^{\vee}= \min(z_a^{\vee}, z_b^{\vee})$. Note that if the intersection is zero (e.g. if $z_a^{\vee} < z_b^{\wedge}$), this will result in $z_{ \cap }^{\wedge} > z_{ \cap }^{\vee}$. Gumbel boxes \cite{dasgupta2020improving} naturally handle these cases and the volume of such ``negative'' boxes is close to zero. 

The conditional probability of one box given another one can be defined as:
\begin{equation}
    P(A|B)= \frac{\mathrm{Vol}(A \cap B)}{\mathrm{Vol}(B)}.
    \label{eq:conditional}
\end{equation}

When using the conditional approach (Section~\ref{sec:approaches}), the values we maximize in \cref{eq:traj} are the conditional probability values obtained in \cref{eq:conditional}. For the symmetric approach, box embeddings offer a variety of possibilities. We list a few of them next ($\mathrm{Sim}$ stands for ``similarity function''):
\begin{itemize}[topsep=0pt,itemsep=-1ex,partopsep=1ex,parsep=1ex,leftmargin=1cm]
\item \textbf{Symmetric conditional}. $\mathrm{Sim} = (P(A | B) + P(B | A)) / 2$
\item \textbf{Intersection over Union (IoU)}. $\mathrm{Sim} = \mathrm{IoU}\rightarrow D = 1 - \mathrm{IoU}= \textrm{Vol}(A \cup B)/\textrm{Vol}(A \cap B)$. This distance is known as the Jaccard distance, and it is a proper metric.
\item \textbf{Sørensen–Dice coefficient}. $\mathrm{Sim} = \textrm{Vol}(A \cup B)/(\textrm{Vol}(A) + \textrm{Vol}(B))$
\item \textbf{Symmetric difference}. $\mathrm{Sim} = \textrm{Vol}(A \triangle B)= \textrm{Vol}(A \cup B)-\textrm{Vol}(A \cap B)$.
\end{itemize}
For the values defined as similarities, we obtain the distance functions (not necessariliy metrics) as $D=1-\mathrm{Sim}$.


\section{Additional Results and Experimental Details}
\label{sec:additional}

\paragraph{Prediction into the future} Our model is better than baselines for every time into the future that we tested. In Figure~\ref{fig:error_time} we show a graph of error for future prediction, with respect to the time elapsed from the end of the past (input) segment. 

\paragraph{Representing multiple futures} In addition to Fig.~\ref{fig:multimodes} in the main paper, in Fig~.\ref{fig:loss_vs_samples}, we show how the evaluation results change depending on how many samples $M$ we use at inference time (the reported value is the best out of the $M$ predictions). Sampling $M>1$ clearly improves the results, which demonstrates that our model represents (and predicts) more than one mode in the distribution. This applies not only to future prediction, but also to past and interpolation prediction.

\paragraph{Error bars} We also report the quantitative results with standard deviations. At test time there are two factors of randomness. First, the input segments can be sampled at different times. Second, there is a sampling process to obtain trajectories $z$ given segment representations $Q$. We run the test with 10 different random seeds, using the same seeds for all the experiments (this is, given a seed, both our method and the baselines use the same sampled segments). In \cref{tab:results} we report the average of the obtained values across the 10 random seeds. In this section we repeat the same results, but add information about the standard deviation across the seeds, shown in parentheses. See \cref{tab:results_std}. Again, we report the $l^2$ error (the lower the better) across keypoints, after normalizing each trajectory to be contained in a region of size $100\times100$. FU, FR, P and I stand for ``future uniform'', ``future random'', ``past'' and ``interpolation'', respectively. The low standard deviation values imply that the significance of the average values reported in \cref{tab:results} is strong. The larger standard deviation values in the long version of the Diving48 dataset reflect the small number of test samples. 

\paragraph{Ablations} We report additional ablations of our framework in Tab.~\ref{tab:ablations}, for the FineGym short dataset. Specifically, we show the following ablations:

\begin{itemize}[topsep=0pt,itemsep=-1ex,partopsep=1ex,parsep=1ex,leftmargin=1cm,labelwidth={1em}]
\item \textbf{No re-encoding}. Same ablation as the main paper one, included here for completeness.
\item \textbf{No trajectory loss}. We train without Eq.~\ref{eq:traj}, which leads to a significant increase in prediction error.
\item \textbf{Gaussian symmetric}. We use Gaussian distributions instead of box embeddings, and train using the symmetric scenario with KL divergence. The conditional case trained with box embeddings (``TrajRep (ours)'' in the table) obtains better results, but that the symmetric case with Gaussian distributions is also competitive, and clearly outperforms the baselines.
\item \textbf{Modify margin $\alpha$}. The choice of $\alpha$ is rather arbitrary. Because the range of the distance function (for the distances used in our experiments) is in $[0, \infty)$, $\alpha$ influences the norm of the distances, but not the relative distances between segments. We run two experiments with different values of $\alpha$ ($0.1$ and $10$), on top of the default $\alpha=1$ and show that the model performance is not very sensitive to this hyperparameter.
\item \textbf{All hard positives}. We replace all soft positives by hard negatives, in order to assess how important is to distinguish them. We notice an increase in error with respect to the original model, showing that distinguishing between hard and soft negatives has some influence in the final model.
\item \textbf{Uniform sampling}. We train using uniform time-steps, and test regularly, with irregular time-steps. This model does not generalize as well to irregular time-steps as the one trained directly on irregular time-steps.
\item \textbf{ST-GCN}. We implement the encoder $\Theta$ using an ST-GCN network \cite{yan2018spatial} instead of a Transformer. ST-GCN is the most established model to process temporal skeleton data. We use the implementation in Yan et. al. \cite{mmskeleton2019}. ST-GCN performs worse than the Transformer network (although the results are competitive), probably because temporal information cannot be added to an out-of-the-box ST-GCN architecture.
\end{itemize}

\begin{table}
\centering
\caption{Ablations on the FineGym short dataset. Values represent mean squared error. See Appendix~\ref{sec:additional} for a discussion.}
\label{tab:ablations}


    \begin{tabular}{l c ccc}
        \toprule

        &&  F      & P       & I  \\
        \midrule
        \textbf{TrajRep (ours)} &       &\ 6.20 & \ 6.36 & \ 4.88\\
        \textbf{\quad w/o re-encoding}     &       &\ 6.49 & \ 6.59 & \ 5.15\\ 
        \textbf{\quad w/o trajectory loss (Eq.~\ref{eq:traj})}     &       & \ 7.37 & \ 7.55 & \ 7.02 \\ 
        \textbf{\quad w/ Gaussian symmetric}     &       & \ 6.64 & \ 6.65 & \ 5.05\\ 
        \textbf{\quad w/ margin $\alpha=0.1$}     &       & \ \textbf{6.18} & \ \textbf{6.32} & \ 4.89 \\ 
        \textbf{\quad w/ margin $\alpha=10$}     &       & \ 6.25 & \ 6.40 & \ \textbf{4.83} \\ 
        \textbf{\quad w/ all hard positives}     &       & \ 6.32 & \ 6.51 & \ 5.03 \\ 
        \textbf{\quad w/ uniform sampling}     &       & \ 6.58 & \ 6.70 & \ 5.36 \\ 
        \textbf{\quad w/ ST-GCN}     &       & \ 6.77 & \ 6.99 & \ 5.26\\ 
    \bottomrule
  \end{tabular}
\end{table}

\begin{figure}
  \centering
  \includegraphics[width=0.7\columnwidth]{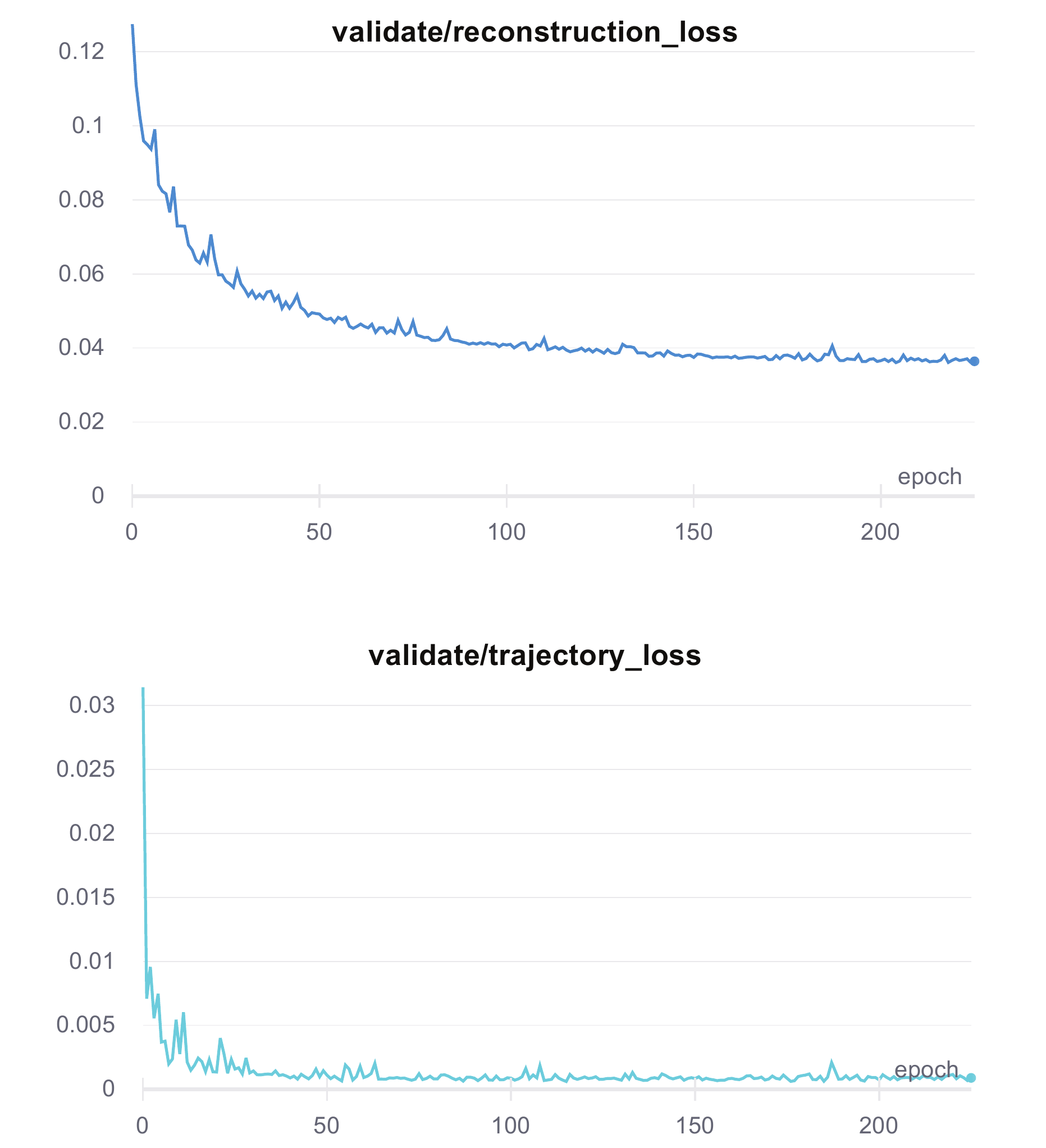}
  \caption{Loss curves during training, corresponding to the FineGym short experiment, trained with bounding boxes.}
  \label{fig:curves}
    \vspace{-0.2cm}
\end{figure}

\paragraph{Stable optimization} The optimization process is stable across the training. We show loss curves in Fig.~\ref{fig:curves} both for $\mathcal{L}_{\mathrm{enc}}$ in Eq.~\ref{eq:traj} (trajectory loss) and for $\mathcal{L}_{\mathrm{dec}}$ in Eq.~\ref{eq:rec} (reconstruction loss).

\paragraph{\cref{fig:editing} details} We find the directions in \cref{fig:editing} by sampling segments from the test set, and for every segment, we modify their time indices in a progressive way, obtaining 10 modified segments for each original one. These modified time-steps are created either by multiplying their time indices by a constant (for Fig.~\ref{fig:speed}), or by adding a constant to their time indices (for Fig.~\ref{fig:offset}). We compute their representations, and for every set of segments we find the main direction of variation across the set by computing the first PCA component. The overall direction is found by averaging these directions across the different test-set samples. This overall direction (for example, the direction representing ``speed change'') is general for all trajectories, not trajectory-specific, and therefore we can apply it to any encoded segment.

\paragraph{Baseline details} Regarding the baselines, the Trajecton++ models the GMM with a first step consisting of a Categorical random variable, followed by a normal distribution. In order to selec the $M=10$ future predictions, we sample the top-10 options in the Categorical variable, and for each one we sample once from the Gaussian. Note that this is exactly how the Trajectron++ is trained, both in the original paper and in our implementation. For the VRNN baseline, at test time we sample the 10 samples in the first time-step, and then sample the mode of the latent distributions for the rest of the samples.

\begin{table}
\centering
\caption{Results from \cref{tab:results} extended with standard deviation information.}
    \label{tab:results_std}

\begin{subtable}{1\linewidth}
\centering

\caption{\textbf{FineGym - Long sequences}}
    \label{tab:long_finegym}
    \begin{tabular}{l c ccc}
        \toprule
        &&  F      & P       & I   \\
            \midrule
\textbf{VRNN} \cite{vrnn}       &       & 15.85 (0.00)& 15.93 (0.00)& 16.10 (0.00)\\
\textbf{Trajectron++ uni.} \cite{salzmann2020trajectron++}      &       & \ 9.54 (0.01)& \ 9.98 (0.01)& \ 9.73 (0.00)\\
\textbf{Trajectron++} \cite{salzmann2020trajectron++}   &       & \ 9.72 (0.00)& 10.01 (0.00)& \ 9.89 (0.00)\\
\textbf{TrajRep (ours, ablation)}     &       & \ 8.82 (0.01)& \ 9.07 (0.00)& \ 7.57 (0.00)\\
\textbf{\quad + re-encoding (ours)} &       & \ 8.50 (0.01)& \ 8.83 (0.00)& \ 7.11 (0.00)\\
        \bottomrule
  \end{tabular}
 \end{subtable}
 
 \begin{subtable}{1\linewidth}
 \centering
    \caption{\textbf{Diving48 - Long sequences}}
    \label{tab:long_diving}
    \begin{tabular}{l c ccc}
        \toprule
        &&  F      & P       & I   \\
            \midrule
\textbf{VRNN} \cite{vrnn}       &       & 23.51 (0.13)& 27.97 (0.17)& 25.66 (0.06)\\
\textbf{Trajectron++ uni.} \cite{salzmann2020trajectron++}      &       & 11.67 (0.22)& 16.52 (0.55)& 11.98 (0.10)\\
\textbf{Trajectron++} \cite{salzmann2020trajectron++}   &       & 11.59 (0.12)& 16.23 (0.21)& 12.68 (0.09)\\
\textbf{TrajRep (ours, ablation)}     &       & 10.00 (0.13)& 11.74 (0.17)& 10.06 (0.13)\\
\textbf{\quad + re-encoding (ours)} &       & \ 9.81 (0.16)& 12.00 (0.11)& \ 9.58 (0.18)\\
        \bottomrule
  \end{tabular}
 \end{subtable}
 
 \begin{subtable}{1\linewidth}
 \centering
    \caption{\textbf{FisV - Long sequences}}
    \label{tab:long_fisv}
    \begin{tabular}{l c ccc}
        \toprule
        &&  F      & P       & I   \\
            \midrule
\textbf{VRNN} \cite{vrnn}       &       & 14.95 (0.00)& 15.03 (0.01)& 15.08 (0.00)\\
\textbf{Trajectron++ uni.} \cite{salzmann2020trajectron++}      &       & 11.42 (0.01)& 11.85 (0.01)& 11.68 (0.01)\\
\textbf{Trajectron++} \cite{salzmann2020trajectron++}   &       & 11.41 (0.00)& 11.71 (0.00)& 11.63 (0.00)\\
\textbf{TrajRep (ours, ablation)}     &       & 10.62 (0.01)& 11.27 (0.01)& \ 9.70 (0.00)\\
\textbf{\quad + re-encoding (ours)} &       & 10.32 (0.01)& 10.77 (0.00)& \ 9.22 (0.00)\\

        \bottomrule
  \end{tabular}
 \end{subtable}
 
 \begin{subtable}{1\linewidth}
 \centering
    \caption{\textbf{FineGym - Short sequences}}
    \label{tab:short_finegym}
    \begin{tabular}{l c ccc}
        \toprule
        &&  F      & P       & I   \\
            \midrule
\textbf{VRNN} \cite{vrnn}       &       & 12.77 (0.00)& 13.20 (0.01)& 13.40 (0.00)\\
\textbf{Trajectron++ uni.} \cite{salzmann2020trajectron++}      &       & \ 7.80 (0.01)& \ 8.28 (0.01)& \ 7.48 (0.01)\\
\textbf{Trajectron++} \cite{salzmann2020trajectron++}   &       & \ 7.26 (0.01)& \ 7.93 (0.01)& \ 6.94 (0.00)\\
\textbf{TrajRep (ours, ablation)}     &       & \ 6.49 (0.00)& \ 6.59 (0.01)& \ 5.15 (0.00)\\
\textbf{\quad + re-encoding (ours)} &       & \ 6.20 (0.01)& \ 6.36 (0.01)& \ 4.88 (0.00)\\
        \bottomrule
  \end{tabular}
 \end{subtable}
 
 \begin{subtable}{1\linewidth}
 \centering
    \caption{\textbf{Diving48 - Short sequences}}
    \label{tab:short_diving}
    \begin{tabular}{l c ccc}
        \toprule
        &&  F      & P       & I   \\
            \midrule
\textbf{VRNN} \cite{vrnn}       &       & 18.36 (0.05)& 20.14 (0.07)& 19.86 (0.02)\\
\textbf{Trajectron++ uni.} \cite{salzmann2020trajectron++}      &       & \ 9.05 (0.02)& 10.36 (0.05)& \ 8.29 (0.03)\\
\textbf{Trajectron++} \cite{salzmann2020trajectron++}   &       & \ 8.74 (0.03)& 11.35 (0.03)& \ 8.31 (0.03)\\
\textbf{TrajRep (ours, ablation)}     &       & \ 6.94 (0.03)& \ 6.99 (0.03)& \ 5.00 (0.02)\\
\textbf{\quad + re-encoding (ours)} &       & \ 6.76 (0.02)& \ 6.85 (0.03)& \ 5.04 (0.02)\\
        \bottomrule
  \end{tabular}
 \end{subtable}
 
 \begin{subtable}{1\linewidth}
 \centering
    \caption{\textbf{FisV - Short sequences}}
    \label{tab:short_fisv}
    \begin{tabular}{l c ccc}
        \toprule
        &&  F      & P       & I   \\
            \midrule
\textbf{VRNN} \cite{vrnn}       &       & 13.26 (0.01)& 13.44 (0.01)& 13.45 (0.00)\\
\textbf{Trajectron++ uni.} \cite{salzmann2020trajectron++}      &       & \ 9.23 (0.01)& \ 9.68 (0.01)& \ 8.86 (0.01)\\
\textbf{Trajectron++} \cite{salzmann2020trajectron++}   &       & \ 8.70 (0.01)& \ 9.28 (0.01)& \ 8.28 (0.00)\\
\textbf{TrajRep (ours, ablation)}     &       & \ 7.83 (0.01)& \ 8.17 (0.01)& \ 6.01 (0.01)\\
\textbf{\quad + re-encoding (ours)} &       & \ 7.54 (0.01)& \ 7.78 (0.01)& \ 5.88 (0.01)\\
        \bottomrule
  \end{tabular}
 \end{subtable}
 \end{table}
 
 \definecolor{mycolor}{HTML}{143a61}
\definecolor{mycolor2}{HTML}{FF5733}
\definecolor{mycolor3}{HTML}{AD23AD}

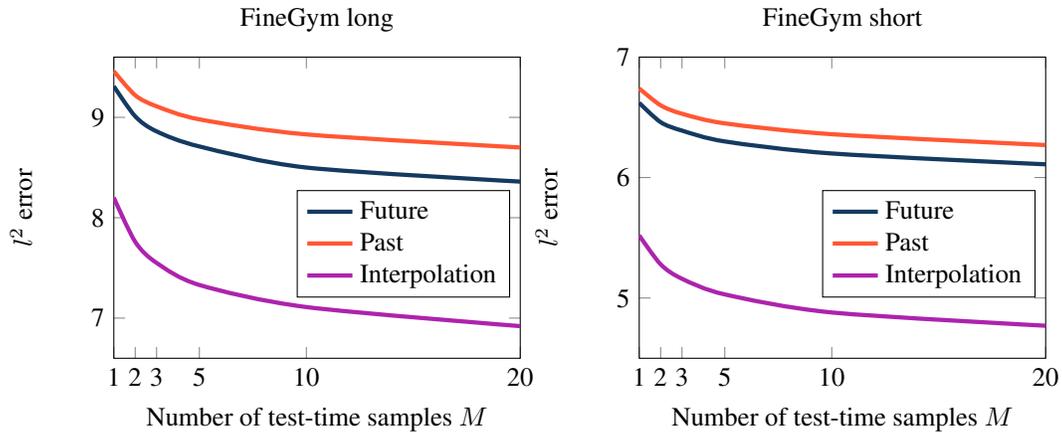
\begin{figure}
    \centering
        \begin{minipage}[t]{.50\textwidth}
        \centering
            \begin{tikzpicture}
                \pgfplotsset{%
                    width=\columnwidth,
                    height=0.8\columnwidth
                }
                \begin{axis}[
                    legend style={at={(0.45,0.38)},anchor=west},
                    legend cell align={left},
                    title=FineGym long,
                    xlabel=Number of test-time samples $M$,
                    ylabel=$l^2$ error,
                    xmin=1, xmax=20,
                    ymin=6.6, ymax=9.6,
                    xtick={1,2,3,5,10,20},
                    xticklabels={1,2,3,5,10,20},   
                    ytick={7, 8, 9},
                    yticklabels={7, 8, 9},
                    every axis plot/.append style={ultra thick}
                            ]
                \addplot[smooth,mycolor] plot coordinates {
                (1, 9.31)
                (2, 9.01)
                (3, 8.86)
                (5, 8.71)
                (10, 8.5)
                (20, 8.36)
                };
                \addlegendentry{Future}
                
                \addplot[smooth,mycolor2] plot coordinates {
                (1, 9.46)
                (2, 9.22)
                (3, 9.11)
                (5, 8.98)
                (10, 8.83)
                (20, 8.7)
                };
                \addlegendentry{Past}
                \addplot[smooth,mycolor3] plot coordinates {
                (1, 8.2)
                (2, 7.76)
                (3, 7.55)
                (5, 7.33)
                (10, 7.11)
                (20, 6.92)
                };
                \addlegendentry{Interpolation}
                \end{axis}
            \end{tikzpicture}
    \end{minipage}\hfill
    \begin{minipage}[t]{.50\textwidth}
        \centering
            \begin{tikzpicture}
                \pgfplotsset{%
                    width=\columnwidth,
                    height=0.8\columnwidth
                }
                \begin{axis}[
                    legend style={at={(0.45,0.38)},anchor=west},
                    legend cell align={left},
                    title=FineGym short,
                    xlabel=Number of test-time samples $M$,
                    ylabel=$l^2$ error,
                    xmin=1, xmax=20,
                    ymin=4.5, ymax=7,
                    xtick={1,2,3,5,10,20},
                    xticklabels={1,2,3,5,10,20},   
                    ytick={5, 6, 7},
                    yticklabels={5, 6, 7},
                    every axis plot/.append style={ultra thick}
                            ]
                \addplot[smooth,mycolor] plot coordinates {
                (1, 6.62)
                (2, 6.46)
                (3, 6.39)
                (5, 6.3)
                (10, 6.2)
                (20, 6.11)
                };
                \addlegendentry{Future}
                
                \addplot[smooth,mycolor2] plot coordinates {
                (1, 6.74)
                (2, 6.6)
                (3, 6.53)
                (5, 6.45)
                (10, 6.36)
                (20, 6.27)
                };
                \addlegendentry{Past}
                \addplot[smooth,mycolor3] plot coordinates {
                (1, 5.52)
                (2, 5.28)
                (3, 5.16)
                (5, 5.03)
                (10, 4.88)
                (20, 4.77)
                };
                \addlegendentry{Interpolation}
                \end{axis}
    \end{tikzpicture}
    \end{minipage}
\caption{In our evaluation, we report the \textit{Best-of-$M$} samples, where $M=10$. In this figure, we show how the evaluation results change for different choices of $M$. These plots show how our model captures different modes in the distribution, as adding more samples results in better values, indicating that the samples result in different trajectories. This is qualitatively reinforced by Fig.~\ref{fig:multimodes}. The figure on the left corresponds to the FineGym long dataset, and the figure on the right corresponds to the FineGym short dataset.} 
\label{fig:loss_vs_samples}
\end{figure}

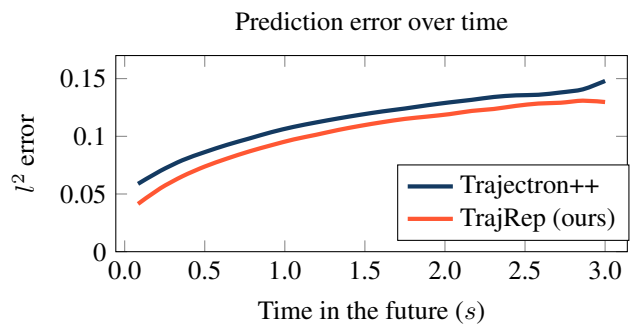
\begin{figure}
\centering
\begin{tikzpicture}
    \pgfplotsset{%
        width=0.6\columnwidth,
        height=0.3\columnwidth
    }
    \begin{axis}[
        title=Prediction error over time,
        legend style={at={(0.55,0.25)},anchor=west},
        legend cell align={left},
        xlabel=Time in the future ($s$),
        ylabel=$l^2$ error,
        xmin=0., xmax=23,
        ymin=0, ymax=0.17,
        xtick={0.4, 4, 7.6, 11.2, 14.8, 18.4, 22},
        xticklabels={0.0, 0.5, 1.0, 1.5, 2.0, 2.5, 3.0},   
        ytick={0,0.05,0.1,0.15},
        yticklabels={0,0.05,0.1,0.15},
        every axis plot/.append style={ultra thick}
                ]
    \addplot[smooth,mycolor] plot coordinates {
    (1, 0.0588)
    (2, 0.07)
    (3, 0.0793)
    (4, 0.0863)
    (5, 0.0926)
    (6, 0.0981)
    (7, 0.1036)
    (8, 0.1082)
    (9, 0.112)
    (10, 0.1155)
    (11, 0.1187)
    (12, 0.1216)
    (13, 0.1241)
    (14, 0.1269)
    (15, 0.1294)
    (16, 0.1316)
    (17, 0.1341)
    (18, 0.1355)
    (19, 0.136)
    (20, 0.138)
    (21, 0.1407)
    (22, 0.148)
    };
    \addlegendentry{Trajectron++}
    
    \addplot[smooth,,mycolor2] plot coordinates {
        (	1	,	0.0414	)
        (	2	,	0.0549	)
        (	3	,	0.0654	)
        (	4	,	0.0738	)
        (	5	,	0.0807	)
        (	6	,	0.0869	)
        (	7	,	0.0923	)
        (	8	,	0.0973	)
        (	9	,	0.1013	)
        (	10	,	0.1055	)
        (	11	,	0.1091	)
        (	12	,	0.1123	)
        (	13	,	0.1151	)
        (	14	,	0.1171	)
        (	15	,	0.1192	)
        (	16	,	0.122	)
        (	17	,	0.1237	)
        (	18	,	0.1264	)
        (	19	,	0.1284	)
        (	20	,	0.1291	)
        (	21	,	0.1309	)
        (	22	,	0.1297	)
    };
    \addlegendentry{TrajRep (ours)}

    \end{axis}
\end{tikzpicture}

\caption{$l^2$ error with respect to the the time elapsed from the past (input) segment, where we evalute future prediction ($t=0$ represents the end of the past segment). This result corresponds to the FineGym long dataset.} 
\label{fig:error_time}
\end{figure}

\end{document}